\documentclass[letterpaper, 10 pt, conference]{ieeeconf}  

\IEEEoverridecommandlockouts                              

\usepackage{graphics} 
\usepackage{amsmath} 
\usepackage{amssymb}  
\usepackage{amstext}
\usepackage{bm}
\usepackage{mathtools}
\usepackage{cite}
\usepackage{xcolor}
\usepackage{glossaries}
\usepackage{url}
\usepackage{breakurl}
\usepackage{hyperref}
\hypersetup{
  bookmarksopen,
  bookmarksnumbered,
  pdfpagemode=UseOutlines,
  colorlinks=true,
  linkcolor=blue,
  anchorcolor=blue,
  citecolor=blue,
  filecolor=blue,
  menucolor=blue,
  urlcolor=blue
}
\usepackage[ruled, noline]{algorithm2e}
\SetKwFor{ForPar}{for}{do in parallel}{end forpar}
\definecolor{alg}{RGB}{46, 149, 186}

\SetCommentSty{mycommfont}
\definecolor{urldarkblue}{RGB}{1, 111, 255}

\usepackage[capitalise]{cleveref} 
\usepackage[font=small]{caption}
\usepackage{multirow}
\usepackage[font=footnotesize]{caption}
\usepackage[labelformat=simple]{subcaption}

\title{\LARGE \bf Learning Implicit Priors for Motion Optimization}

\author{Julen Urain$^{2\,*}$, An T. Le$^{2\,*}$, Alexander Lambert$^{1\,*}$, Georgia Chalvatzaki$^{2}$, Byron Boots$^{1}$ and, Jan Peters$^{2}$
\thanks{$*$ All authors contributed equally.}
\thanks{$^{1}$ Paul G. Allen School of Computer Science and Engineering (CSE), University of Washington (USA)}%
\thanks{$^{2}$ Intelligent Autonomous Systems Lab,
        Technische Universität Darmstadt (Germany)}%
\thanks{This project has received funding from the European Unions Horizon 2020 research and innovation programme under grant agreement No \# 820807 (SHAREWORK) and DFG project, PE 2315/11-1 (METRIC4IMITATION)}%
}

\newacronym{ebm}{EBM}{Energy Based Models}
\newacronym{nce}{NCE}{Noise Contrastive Estimation}

\newacronym{promp}{ProMP}{Probabilistic Movement Primitives}
\newacronym{gp}{GP}{Gaussian Process}
\newacronym{vae}{VAE}{Variational Autoencoder}
\newacronym{gan}{GAN}{Generative Adversarial Network}
\newacronym{chomp}{CHOMP}{Covariant Hamiltonian Optimization for Motion Planning}
\newacronym{stomp}{STOMP}{Stochastic Trajectory Optimization for Motion Planning}
\newacronym{gpmp}{GPMP}{Gaussian Process Motion Planning}
\newacronym{sgpmp}{StochGPMP}{Stochastic Gaussian Process Motion Planning}

\newacronym{map}{MAP}{\textit{maximum a-posteriori}}

\newacronym{cd}{CD}{Contrastive Divergence}

\newacronym{mle}{MLE}{Maximum Likelihood Estimation}
\newacronym{mcmc}{MCMC}{Markov Chain Monte Carlo}

\newacronym{ioc}{IOC}{Inverse Optimal Control}
\newacronym{irl}{IRL}{Inverse Reinforcement Learning}

\def\1{\bm{1}}

\def\RR{\mathbb{R}}

\def\d{\text{d}}

\def\vmu{{\bm{\mu}}}
\def\vtheta{{\bm{\theta}}}

\def\vf{{\bm{f}}}

\def\vq{{\bm{q}}}

\def\vv{{\bm{v}}}

\def\vx{{\bm{x}}}

\def\vepsilon{{\boldsymbol{\epsilon}}}

\def\vmu{{\boldsymbol{\mu}}}

\def\vsigma{{\boldsymbol{\sigma}}}
\def\vtheta{{\boldsymbol{\theta}}}
\def\vtau{{\boldsymbol{\tau}}}

\def\vSigma{{\boldsymbol{\Sigma}}}

\def\mI{{\bm{I}}}

\DeclareMathAlphabet{\mathsfit}{\encodingdefault}{\sfdefault}{m}{sl}
\SetMathAlphabet{\mathsfit}{bold}{\encodingdefault}{\sfdefault}{bx}{n}

\newcommand{\expect}[2]{\mathbb{E}_{#1}\Big[ #2 \Big]}

\def\gD{{\mathcal{D}}}
\def\gE{{\mathcal{E}}}

\def\gL{{\mathcal{L}}}

\def\gN{{\mathcal{N}}}

\def\gX{{\mathcal{X}}}

\newcommand{\E}{\mathbb{E}}

\DeclareMathOperator*{\argmin}{arg\,min}

\makeatletter
\newcommand{\StatexIndent}[1][3]{%
  \setlength\@tempdima{\algorithmicindent}%
  \Statex\hskip\dimexpr#1\@tempdima\relax}
\makeatother

\newcommand{\ie}{\textit{i.e.}}

\def\mub{\bm{\mu}}
\def\mutens{\bm{M}}
\def\tb{\textbf{\textit{t}}}
\def\Kb{\bm{\mathcal{K}}}

\def\J\mathbf{J}

\def\thetab{\mathrm{\boldsymbol{\theta}}}
\def\stateb{\bm{x}}

\DeclarePairedDelimiterX{\infdivx}[2]{(}{)}{%
  #1\;\delimsize\|\;#2%
}

\newcommand{\kldiv}{\mathrm{KL\,}\infdivx}

\newcommand{\hide}[1]{}
\newcommand{\norm}[1]{\big|\big| #1 \big|\big|}

\begin{document}

\maketitle
\thispagestyle{empty}
\pagestyle{empty}

\begin{abstract}
Motion optimization is an effective framework for generating smooth and safe trajectories for robotic manipulation tasks. However, it suffers from local optima that hinder its applicability, especially for multi-objective tasks. In this paper, we study this problem in light of the integration of Energy-Based Models (EBM) as guiding priors in motion optimization. EBMs are probabilistic models with unnormalized energy functions that represent expressive multimodal distributions. Due to their implicit nature, EBMs can easily be integrated as data-driven factors or initial sampling distributions in the motion optimization problem. This work presents a set of necessary modeling and algorithmic choices to effectively learn and integrate EBMs into motion optimization. We present a set of EBM architectures for learning generalizable distributions over trajectories that are important for the subsequent deployment of EBMs. Moreover, we investigate the benefit of including smoothness regularization in the learning process to improve motion optimization. In addition to gradient-based solvers, we also propose a stochastic method for trajectory optimization with learned EBMs. We provide extensive empirical results in a set of representative tasks against competitive baselines that demonstrate the superiority of EBMs as priors in motion optimization scaling up to 7-dof robot pouring that can be easily transferred to the real robotic system. Videos and additional details are available at \url{https://sites.google.com/view/implicit-priors}.
\end{abstract}

\section{Introduction}
\textit{Motion planning} is a fundamental property for autonomous robots to achieve task-specific goals. In the context of autonomous robot manipulation, the trajectories that the robot should execute should satisfy several constraints, e.g., approaching the goal while avoiding collisions with itself and the world and joint limits. Naturally, a complex motion plan can be viewed as a multi-objective optimization problem along a specific time horizon. In this work, we study motion planning in light of \textit{motion optimization} methods, when multiple objectives need to be satisfied for accomplishing an end task. In contrast with sampling-based motion planning methods~\cite{kuffner2000rrt, kavraki1996probabilistic}, motion optimization methods cast the motion planning problem into a trajectory optimization one~\cite{ratliff2009chomp, kalakrishnan2011stomp, schulman2014motion, mukadam2018continuous}, which optimize an initial trajectory or trajectory distribution iteratively by gradient descent or stochastic optimization until convergence. 

Motion optimization is an inherently local optimization method that relies on the initialization, iteratively making local updates at every optimization step. Hence, due to the possible non-convexity of the cost function, these optimization methods suffer from local minima. Additionally, if the cost function is sparse, it might be hard to get the proper information for reaching low-cost regions,  and therefore, the initial proposal may barely improve.

\begin{figure}[!t]
  \begin{subfigure}[b]{0.4\linewidth}
    \centering
    \includegraphics[height=8.5\baselineskip]{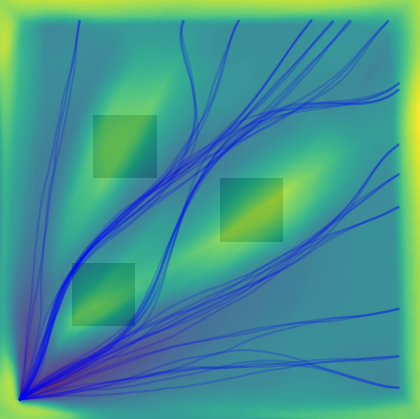}
    \caption{EBM for trajectories}
    \label{subfig:ebm_trajopt}
  \end{subfigure}
  \hspace{-11pt}
  \begin{subfigure}[b]{0.6\linewidth}
    \centering
    \includegraphics[height=8.5\baselineskip]{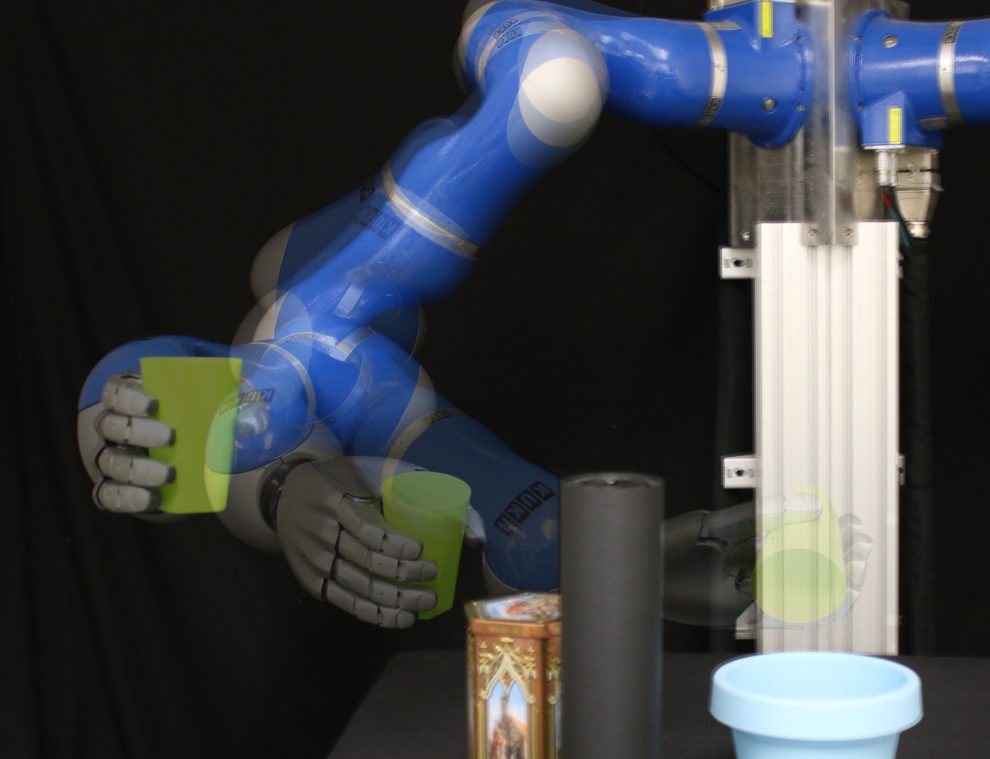}
    \caption{Pouring Task}
    \label{subfig:pouring}
  \end{subfigure}
\caption{Implicit trajectory distributions are learned from demonstrations using EBMs. These guide motion optimization to produce feasible trajectories for new problems. \subref{subfig:ebm_trajopt} An obstacle avoidance energy function, with generated optimal trajectories towards different goal locations. \subref{subfig:pouring} A robot manipulator using learned EBMs to pour a cup within a cluttered scene.}\vspace{-18pt}
\label{fig:main_fig}
\end{figure}

A way to avoid the local minima traps in trajectory optimization is to include a set of informative priors in the trajectory optimization. 
Classical motion optimization methods usually employ a set of handcrafted initialization proposals to warm-start the optimization. In \gls{chomp}~\cite{ratliff2009chomp}, the initial trajectory is a straight line that moves from the initial to the goal configuration. 
\gls{stomp}~\cite{kalakrishnan2011stomp}, on the other hand, considers a normal initial sampling  distribution with a structured covariance matrix modeled to increase entropy in the center of the state-based trajectory, with minimum variance at the start- and end-points. Alternatively, engineered costs might help guide or regularize the optimization. However, these handcrafted heuristics can be difficult to tune correctly in practice, and can do little to alleviate multimodality of complex manipulation tasks.

In a different vein, to capture the inherent multimodality of multi-objective motion planning tasks, a line of work proposes to learn trajectory distributions from datato guide the optimization process away from local minima. Explicitly, these methods fit a density model to a set of provided demonstrated trajectories~\cite{urain2020imitationflow} and then use these models as the guiding priors for motion optimization \cite{koert2016demonstration, pmlr-v78-rana17a}. However, these approaches often struggle in capturing multi-modality in high-dimensional spaces, leading to sub-optimal solutions. 


In this work, we study the modeling of priors for motion optimization as \textit{implicit models}~\cite{du2019implicit}. Namely, we propose to use \gls{ebm}~\cite{lecun2006tutorial} as our density models, a class of implicit functions, that can represent un-normalized density functions that can easily capture multi-modalities in data. \gls{ebm}s can be represented in arbitrary latent spaces and then, transform the distribution to a higher dimensional space~\cite{urain2021composable}. This property allows us to represent configuration space trajectory distributions given an \gls{ebm} in a lower-dimensional task space. Additionally, due to their exponential nature, \gls{ebm}s can be easily combined, allowing the composition of multiple priors, representing sub-tasks of the manipulation task, into a single structured prior.

We propose a motion optimization framework using as priors implicit functions that are modular, learnable, differentiable, and composable. Using our learned \gls{ebm}s as priors, we can integrate multimodal information that can bias and guide the optimization process towards finding a feasible and smooth solution in complex tasks. Concretely, in this paper, we \textit{a)} show how training with smoothness regularization and task-specific \gls{ebm} design choices such as object-centric or phase conditioning can benefit motion optimization, \textit{b)} explain how \gls{ebm}s can naturally be integrated as structured priors in a motion optimization problem,  \textit{c)} propose a novel stochastic trajectory optimization method using \gls{gp} priors, and, \textit{d)} demonstrate the applicability and effectiveness of our motion optimization framework using \gls{ebm}s as structured priors to guide the optimization problem both in simulation and real-world robotic tasks.

\section{Related Work}
\noindent\textbf{Motion optimization.} 
While sampling-based motion planning algorithms have gained significant traction \cite{kavraki1996probabilistic,kuffner2000rrt}, they are typically computationally expensive, hindering their application in real-world problems. Moreover, these methods cannot guarantee smoothness in the trajectory execution, resulting in jerky robot motions that must be post-processed before applying them on a high degree-of-freedom (DoF) robot. Contrarily to sampling-based motion planners, trajectory optimization methods can integrate collisions and motion smoothness constraints, optimizing over an initially proposed trajectory based on a set of cost functions. \gls{chomp} and its variants \cite{he2013multigrid,byravan2014space,marinho2016functional} optimize a cost function using covariant gradient descent over an initial suboptimal trajectory that connects the starting and goal configuration. \gls{stomp} proposed to optimize over non-differentiable constraints by drawing stochastic samples from a set of noisy trajectories, making \gls{stomp} increasingly dependent on the parameters of the noisy distribution. Due to their assumptions, these methods may not find an optimal solution or even fail to converge. TrajOpt \cite{schulman2014motion} addresses the computational complexity of \gls{chomp} and \gls{stomp}, that require a fine discretization of the trajectory for collision checking, proposing a sequential quadratic program with continuous collision-time collision checking. In \gls{gpmp}~\cite{mukadam2018continuous}, motion optimization is cast as a probabilistic inference problem. A trajectory is parameterized as a function of continuous-time that maps to robot states, while a \gls{gp} is used as a prior distribution to encourage trajectory smoothness, while a likelihood function encodes feasibility. The trajectory is calculated via \gls{map} estimation from the posterior distribution of trajectories from the \gls{gp} prior and the likelihood function.

\noindent\textbf{Learning for motion generation.} Learning-based methods for robot motion generation hold the promise of generalizing robot skills \cite{kober2013reinforcement}. Reinforcement learning provides reactive policies that usually overfit a single task \cite{ibarz2021train}, while imitation learning aims to provide policies learned through demonstrations \cite{osa2018algorithmic}. For the latter, behavioral cloning (BC) methods try to explicitly replicate the demonstrated trajectory \cite{torabi2018behavioral, zhang2018deep}, but the acquired policies do not extrapolate outside the data distribution, are usually unimodal, and tend to fail when task conditions change. However, these explicit models have been integrated into motion optimization frameworks as a way to guide the optimization process, to tackle issues of computational complexity and trapping to local minima \cite{koert2016demonstration,pmlr-v78-rana17a}. Recently, various works explored the integration of learned priors for motion planning \cite{ichter2018learning,ortiz2022structured, chamzas2021learning,qureshi2020motion,cheng2020learning}. Unfortunately, the explicit models representing the data distribution are prone to collapse into single modes and do not adequately capture the multimodality of complex manipulation tasks; therefore, their contribution is limited to simple use-cases of robot motion generation. Hence, implicit models seem a more viable option \cite{zhang2018deep,kingston2019exploring}.

Highly connected to our work are the fields of \gls{ioc}~\cite{levine2012continuous, finn2016guided} and \gls{irl}~\cite{ng2000algorithms, abbeel2004apprenticeship,ziebart2008maximum} methods. In \gls{ioc} and \gls{irl}, given a set of demonstrations, we aim to learn the cost function that these demonstrations are trying to maximize. In particular, Maximum Entropy IRL (MaxEntIRL) \cite{ziebart2008maximum} models the demonstrations as Gibbs distribution, whose energy is given by the unknown cost function, i.e., it approximates an \gls{ebm}.
Although our work draws connections with MaxEntIRL, and investigates the learning of implicit models for robotic manipulations tasks, our contributions are distinctive. Crucially, we study \gls{ebm}s as priors for motion optimization, proposing targeted ways for learning and integrating \gls{ebm}s in robot motion optimization.



\section{Background}

\noindent\textbf{Energy based models} are probabilistic models that represent a probability density function $p_{\vtheta}$ up to an unknown normalizing constant,
\begin{align}
    p_{\vtheta}(\vx) = \frac{1}{Z_{\vtheta}}\exp(-E_{\vtheta}(\vx)),
\end{align}
\noindent with $\vx \in \gX$ the input variable, $E_{\vtheta}: \gX \xrightarrow{} \RR$ the energy function and $Z_\vtheta =\int_{\gX} \exp(- E_{\vtheta}(\vx))\d \vx$ the normalization constant. In an \gls{ebm}, the energy function $E_{\vtheta}$ is usually modelled by an arbitrary neural network parameterized by $\vtheta$. Due to the arbitrary shape of the energy function $E_{\vtheta}$, the computation of $Z_{\vtheta}$ is usually intractable.

\noindent{\textit{Training of \gls{ebm}s.}} Even if \gls{ebm}s have been widely applied in a different set of applications, one of the most common applications is to use them as generative models.
Given a dataset $\gD:\{\vx_d\}_{0:N}$, we aim to fit our parameterized density model to a dataset.
There exist several algorithms for learning \cite{song2021train}. In our work, we use \gls{cd}~\cite{hinton2002training}. In \gls{cd}, the \gls{mle} objective is re-written to maximize the probability of the samples from the data compared to randomly sampled \textit{negative} data points,
\begin{align}
    \label{eq:cd}
    \gL_{\textrm{CD}} = \E_{p_{\gD}(\vx)}\left[ E_{\vtheta}(\vx) \right] - \E_{q(\vx)}\left[E_{\vtheta}(\vx) \right],
\end{align}
with $p_{\gD}$ being the distribution of the data and $q$ the distribution of the negative samples.
Optimally, the negative samples' distribution should match the \gls{ebm}, $q(\vx) \propto \exp (-E_{\vtheta}(\vx))$. Nevertheless, in practice multiple different approaches are considered to sample the negative samples, such as uniform sampling, Langevin \gls{mcmc}~\cite{parisi1981correlation} over the \gls{ebm}~\cite{du2019implicit}, or learning an approximated sampler~\cite{kumar2019maximum, grathwohl2020nomcmc}. It is a good practice to adapt the negative sampling distribution to the problem domain.

\noindent\textit{Sampling from the \gls{ebm}.} In contrast with most generative models~\cite{goodfellow2014generative, DBLP:journals/corr/KingmaW13}, given the implicit nature of \gls{ebm}s, sampling from an \gls{ebm} cannot be done explicitly. Instead, there exist multiple algorithms to generate samples from the learned energy function; from Rejection Sampling~\cite{gilks1992adaptive} to \gls{mcmc}~\cite{andrieu2003introduction} with the most popular being sampling with Langevin Dynamics. 

\noindent\textbf{Trajectory optimization.} Denoting the system state at time $t$ to be $\vx_t\in \mathbb{R}^d$, we can define a discrete-time trajectory as the sequence $\vtau \triangleq (\vx_0, \vx_1, ..., \vx_{T-1}, \vx_T)$ over a planning horizon $T$. For a given start-state $\vx_0$, trajectory optimization aims to find the optimal trajectory $\vtau^*$ which minimizes an objective function $c(\vtau, \vx_0)$. This function might include a cost $c_g$ on distance to a desired goal-state, $\vx_g$, along with a cost $c_{obs}$ on collisions to promote obstacle avoidance. An additional term is often incorporated to penalize non-smooth trajectories, which we denote as $c_{sm}$. The overall objective can then be expressed as the sum of these individual terms
\begin{align}
    \label{eq:traj_opt}
    \vtau^* &= \arg \min_{\vtau}c(\vtau, \vx_0, \vx_g) \\
    &= \arg \min_{\vtau} c_{obs}(\vtau) +  c_g(\vtau, \vx_g) + c_{sm}(\vtau, \vx_0).
\end{align}
Summarizing the \textit{context} parameters of the planning problem as $\mathcal{E} = [\vx_0, \vx_g, ...]^\top$, the objective function can also be written more generally to include any number of cost terms: $c(\vtau, \mathcal{E})=\sum_i c_i(\vtau, \mathcal{E})$.

Gradient-based strategies for trajectory optimization typically resort to second-order iterative methods similar to Gauss-Newton~\cite{toussaint2014newton, mukadam2018continuous}, or use pre-conditioned gradient-descent~\cite{ratliff2009chomp} to find a locally optimal solution to the objective. However, this requires that the cost function be once- or twice-differentiable, leading to carefully handcrafted cost terms such as truncated signed-distance fields (t-SDF), which must be pre-computed for a given environment.

On the other hand, sampling-based approaches resort to stochastics generation of candidate trajectories using a proposal distribution. These samples are then evaluated on the objective and weighted according to their relative performance~\cite{kalakrishnan2011stomp, botev2013cross}. Such gradient-free optimization can operate on discontinuous costs (which may arise from contact between surfaces, for example). However, the inherent stochasticity may lead to undesirable oscillatory behavior and require additional heuristics to achieve satisfactory performance~\cite{bhardwaj2022storm}.

\noindent\textbf{Planning as inference.}
The duality between probabilistic inference and optimization for planning and control has been widely explored~\cite{toussaint2009robot, levine2018reinforcement, mukadam2018continuous}. With open-loop trajectory optimization, in particular, we view the trajectory $\vtau$ as a random variable and first consider the target distribution
\begin{align}\label{eq:map_posterior}
    p(\vtau; \mathcal{E}) = \frac{1}{Z}\, \prod_i p_i(\vtau; \mathcal{E}),
\end{align}
where each $p_i$ term consists of an individual probability factor (which can also be a prior term on $\vtau$). Optimization can then be formulated as a \textit{maximum a posteriori} (MAP) problem, where we seek to find $\vtau^*=\arg\max_{\vtau}p(\vtau;\mathcal{E})$. This can be done by minimizing the negative-log of the distribution
\begin{align}\label{eq:map}
    \vtau^*= \vtau_{\textrm{MAP}} &= \arg \min_{\vtau}-\log \prod_i p_i(\vtau; \mathcal{E}).
\end{align}
Assuming that these probability densities belong to the exponential family, we can relate them to the previous cost terms: $p_i(\vtau; \mathcal{E}) \propto \exp(-c_i(\vtau; \mathcal{E}))$. Substituting these into \cref{eq:map} recovers \cref{eq:traj_opt}. 
This perspective is important, as it justifies the use of optimization methods for inference, and permits the natural integration of trajectory distributions in the optimization problem.

\section{Learning \gls{ebm}s for motion optimization}\label{sec:learn_ebm}



Given a new context $\gE$, performing inference over the posterior distribution in \cref{eq:map_posterior} and \cref{eq:map} requires that we define a prior distribution of trajectories, $p(\vtau; \mathcal{E})$.  Given a dataset $\gD=\{\vtau_j, \gE_j\}_{j=1:N}$, we propose to model and \textit{learn} such prior trajectory distributions from collected data. We define this distribution as an EBM,
\begin{align}\label{eq:ebm_prior}
    p_\vtheta(\vtau | \gD; \gE) = \frac{1}{Z} \exp (-E_{\vtheta}(\vtau, \gE)) .
\end{align}
with model parameters $\vtheta$. In practice, $\gE$ represents the planning contexts, e.g., goal targets, obstacles positions, trajectory phase, etc.
The dataset $\gD$ may consist of a collection of expert demonstrations on different environments,
and we aim to fit a density function representing the data distribution with \gls{cd}~\cref{eq:cd}.  While the learned prior distribution is based on a set of demonstrations, we desire to adapt to novel scenarios beyond the demonstrated examples. The learned distribution is expected to provide informative samples beyond the demonstrated cases.
Notably, instead of learning a monolithic \gls{ebm}-based prior, \textsl{we can factor this prior distribution depending on various aspects of the problem}
\begin{align}\label{eq:ebm_factored2}
    p_\vtheta(\vtau | \gD; \gE) \propto \prod_i\exp (-E_{\vtheta_i}(\vtau, \gE_i)).
\end{align}
Such a factored distribution allows us to leverage composability, learn modular \gls{ebm} factors independently, and combine them as needed for novel scenarios and planning problems~\cite{urain2021composable}.
Furthermore, to properly learn \gls{ebm}s for motion optimization, we need to make multiple algorithmic and modeling choices. In the following, we introduce a set of proposed choices to properly learn and represent high-dimensional, long-horizon multimodal trajectory distributions via \gls{ebm}s that can be beneficial for their deployment in motion optimization.





\begin{figure}
    \centering
    \begin{minipage}{.22\textwidth}
		\centering
		\includegraphics[width=.95\textwidth]{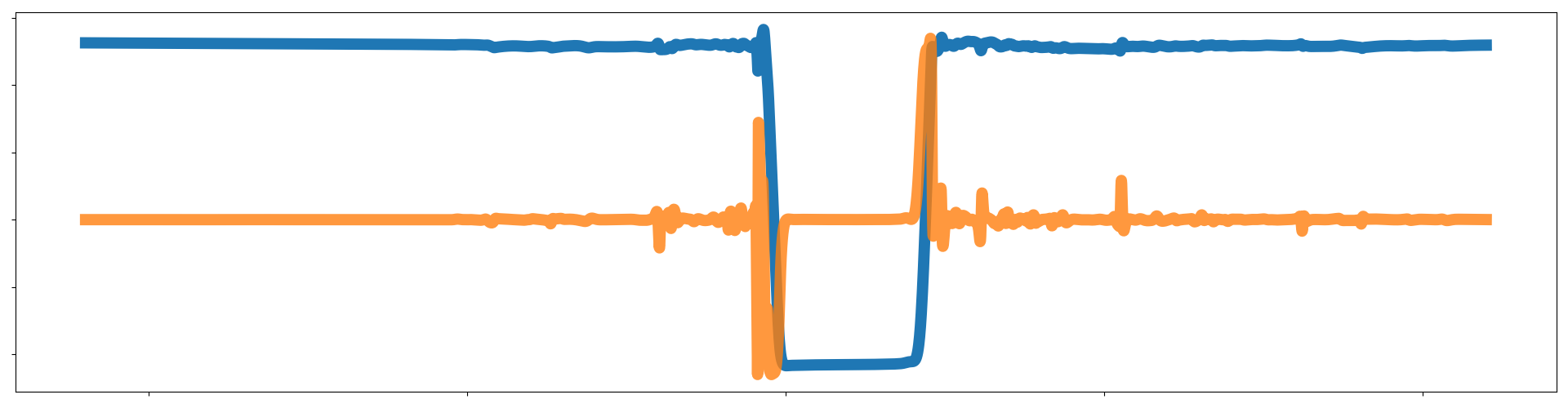}
	\end{minipage}
	\begin{minipage}{.22\textwidth}
		\centering
		\includegraphics[width=.95\textwidth]{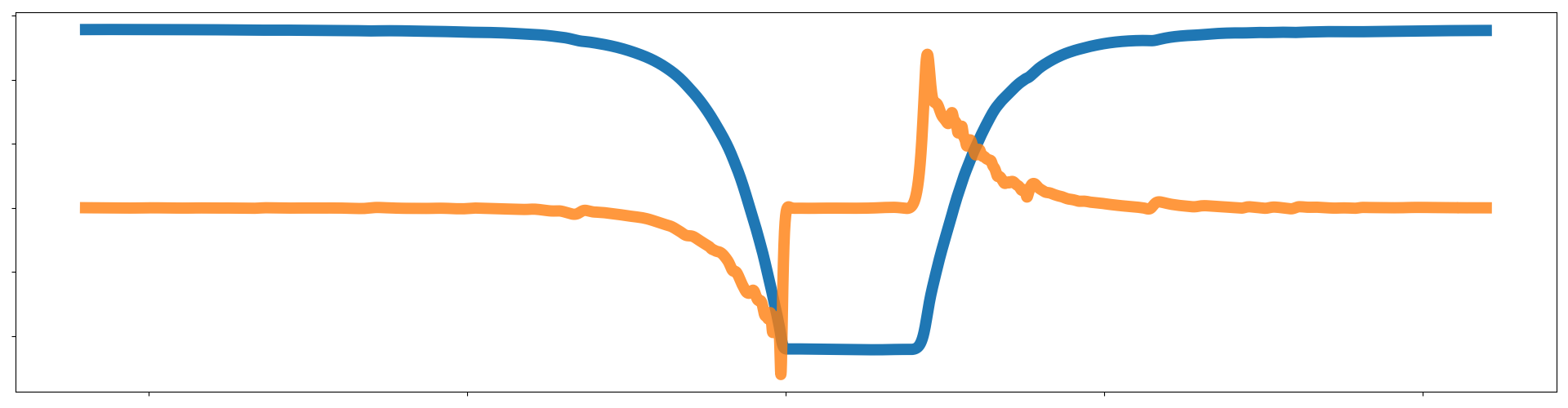}
	\end{minipage}
    \caption{Learned \gls{ebm}~(blue) and its gradient~(orange) in a 1D dataset. (Left) \gls{ebm} trained with vanilla \gls{cd}. (Right) \gls{ebm} trained with \gls{cd} loss + denoising regularizer~\eqref{eq:dsm}.}
    \label{fig:ebm_landscape}
    \vspace{-15.5pt}
\end{figure}

\noindent\textbf{Smoothing \gls{ebm}s for gradient-based optimization.}
To deploy \gls{ebm}s in motion optimization, we need a smooth energy landscape. However, the \gls{cd} objective of \cref{eq:cd} generates an energy landscape with multiple plateaus, with high energy values in regions where there are no data and a plateau of low-energy in the regions of the data points. While the energy landscape may capture well the distribution of the demonstrations, it might not be helpful for gradient-based sampling or optimization, with gradients close to zero in the plateau and high gradients in the cliffs (\cref{fig:ebm_landscape}). We propose adding a denoising score matching loss~\cite{vincent2011connection, song2021scorebased} as regularization to smoothen the energy landscape and improve gradient information. Denoising score matching first generates a noisy sample given a data sample $\Tilde{\vx}\sim p(\Tilde{\vx}|\vx)= \gN(\Tilde{\vx}|\vx, \vsigma^2 \mI)$, as $  \Tilde{\vx} = \vx + \vsigma\vepsilon$, with $\vepsilon \sim \gN(0, \mI)$, and, then, matches the score function, $\nabla_{\vx}E_{\vtheta}$ to denoise the sample $\Tilde{\vx}$ back to $\vx$ 
\begin{align}
    \label{eq:dsm}
    \gL_{\textrm{DSM}} = \E_{p_{\gD}(\vx, \gE)}\E_{p(\Tilde{\vx}|\vx)}\left[ ||\vepsilon -\nabla_{\vx}E_{\vtheta}(\Tilde{\vx}, \gE) ||_2^2 \right].
\end{align}
The loss \eqref{eq:dsm} encourages the gradient of the \gls{ebm} to point towards the data distribution contrarily to the \gls{cd} loss. 

\noindent\textbf{Task-specific \gls{ebm}s for Motion Optimization}
Here, we introduce a set of model choices to represent \gls{ebm}s for motion optimization,
as making proper choices on the \gls{ebm} architecture improves the data representation capacity and the generalization of the learned models. 

\noindent\textit{{Object-centric \gls{ebm}s.}} 
Learning task-conditioned motion models is a vital tool for representing task-adaptive motion behaviors. In our work, we propose to learn object-centric \gls{ebm}  that are useful for representing desired movements in manipulation tasks that involve objects, conditioning the learned \gls{ebm} on the objects' poses.



\noindent\textit{{Phase-conditioned \gls{ebm}.}}
Learning density models directly on the trajectory level requires modeling in a $T\times D$ dimensional space, with $T$ temporal horizon and $D$-dimensional state-space. Learning phase-conditioned density models is particularly challenging, as with long-horizon trajectories, the dimension of the input space increases, and the learning of an \gls{ebm} in that space might be challenging. Additionally, $T$ forces the temporal horizon to be discretized with a specific frequency, and it might be hard to use it for different discretization frequencies or even to represent continuous-time distributions. The usability of phase-conditioned priors for motion optimization is necessary for long-horizon tasks. Therefore, we need to learn phase-conditioned \gls{ebm}s
\begin{align}
    p(\vx| \alpha) \propto \exp(-E_{\vtheta}(\vx, \alpha)),
\end{align}
with $\vx$ the state and $\alpha$ being the phase.
The phase represents a continuous variable moving from $0$ to $1$, encoding the temporal evolution of the manipulation task. The phase-conditioned \gls{ebm} represents the state-occupancy distribution for different instances of the manipulation task. Nevertheless, the phase-conditioned \gls{ebm} lacks any temporal relation between temporally adjacent points, generating non-smooth trajectories.  To confront this effect, we propose combining phase-conditioned \gls{ebm}s with trajectory smoothing costs to represent smooth trajectory distributions
\begin{align}
    \label{eq:phase_ebm_trj}
    p(\vtau) \propto \exp \left(-\sum_{k} E_{\vtheta}(\vx_k,\alpha_k)  + (\vx_k - \vx_{k+1})^2 \right).
\end{align}

\section{Combining \gls{ebm}s and motion optimization}\label{sec:ebm_mo}

Although we may be able to learn data-driven prior distributions, as described in the previous section, we still require reliable inference and optimization methods to derive optimal trajectories $\vtau^*$ given a new planning problem expressed by $\gE$.
In the following section, we focus on methods for evaluation and inference on learned EBMs for trajectory distributions. This includes techniques for sampling and optimization which account for the compose-ability of our EBM functions, as well as a stochastic trajectory optimization method suited for the planning tasks considered here.

\noindent\textbf{Structured Planning Priors.}
Since we cannot sample from EBMs directly, we need to initialize samples by first drawing from an initial distribution, and then perform sequential updates to approximate the relevant modes. Further, learning task-specific EBM model-components is useful for portions of the objective function which are hard to define. However, we may insist on biasing our sampling given the structure of the planning problem, and incorporate known, well-defined requirements such as goal-seeking behavior and smoothness. This can be addressed by incorporating relevant distributions which are known \textit{a-priori}, to the contextual-prior in \cref{eq:ebm_prior}
\begin{align}\label{eq:ebm_factored}
    p_\vtheta(\vtau | \gD; \gE) \propto p_0(\vtau; \gE) \prod_i\exp (-E_{\vtheta_i}(\vtau, \gE_i)),
\end{align}
where $p_0(\vtau; \gE)$ is a general trajectory-based prior. Similarly to \cite{mukadam2018continuous, lambert2021entropy}, we can directly integrate goal-reaching and smoothness into this distribution, which can then be directly sampled to initialize optimization. 
This distribution is composed using a state-based start prior  $p_s(\vx)=\mathcal{N}(\vmu_s, \vSigma_s)$, goal-directed prior $p_g(\vx)=\mathcal{N}(\vmu_g, \vSigma_g)$, where $\vmu_s$, $\vmu_g$ are the expected start and goal configurations, respectively. Smoothness is defined using a time-correlated Gaussian-Process prior $p^{gp}$ derived from a linear time varying motion model (see Appendix A in \cite{alwala2020joint} for details).
These terms can then be composed
\begin{align}
    p_0(\vtau; \gE) \, &= \, p_s(\vx_0)\  p_g(\vx_T)  \prod_{t=0}^{T-1} p_t^{gp} (\vx_t, \vx_{t+1})\\
    & \,\propto \exp\big(-\frac{1}{2}\big|\big| \vtau - \vmu_\vtau\big|\big|^2_{\mathcal{K}}\big)
\end{align}
where $\gE = \{\vmu_s, \vmu_g\}$, $\mathcal{K}$ is a block-diagonal covariance matrix, and $\vmu_\tau$ the straight-line trajectory defined in configuration space. This results in a closed-form, explicit distribution with low entropy near start and goal regions, and high-entropy towards the middle of a trajectory, which is ideal for goal-centric planning tasks. In contrast to \cite{kalakrishnan2011stomp}, the integrated dynamics includes higher-order terms, such that velocities and accelerations can also be sampled in a principled manner. In practice, we can efficiently generate large quantities of these time-correlated trajectories due to our parallelized GPU implementation. Note, however, it is intractable to incorporate explicit priors on behaviors such as obstacle avoidance, e.g. configuration space obstacle avoidance prior. Hence, in practice we must resort to a \textit{combination} of implicit and explicit priors to generate feasible trajectories from \cref{eq:ebm_factored}.

\noindent\textbf{Stochastic Trajectory Optimization with GP-Priors.}
We can iteratively update the time-correlated sampling prior, described in the previous section, to fit the modes of a learned EBM and allow us to sample optimal trajectories in a new planning context. The optimizer, which we call \gls{sgpmp}, is closely related to the importance sampling scheme used by CEM and MPPI but uses goal-directed GP distributions. We refer the reader to the Appendix on the \href{https://sites.google.com/view/implicit-priors}{project site} for details on the derivation and specific update procedure. When used in our experiments, we select to update the mean of each component GP in the distribution. This can be easily performed in configuration space, although it requires an initial goal configuration to be approximated using Inverse-Kinematics.
 

\label{sec:grad_descent_optimization}

\section{Experimental Evaluation}
\label{sec:experiments}



\begin{figure}[!t]
	\centering
	\begin{minipage}{.49\columnwidth}
		\centering
		\includegraphics[width=.95\textwidth]{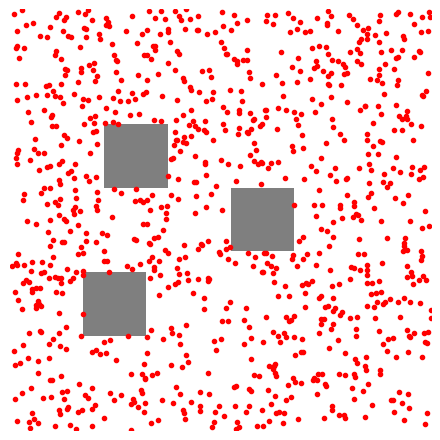}
	\end{minipage}
	\begin{minipage}{.49\columnwidth}
		\centering
		\includegraphics[width=.95\textwidth]{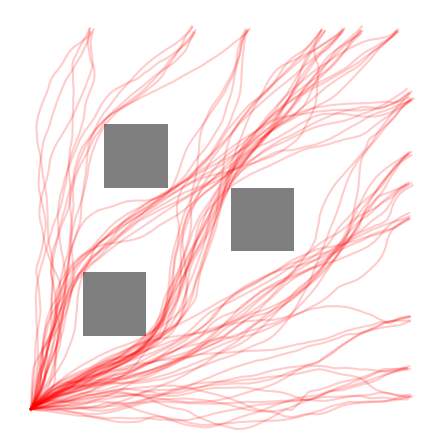}
	\end{minipage}
	\caption{Examples of positive sampling distributions for a single obstacle environment. (Left) Uniform point-based sampling with rejection. (Right) Expert trajectory distributions for randomly sampled goal locations along the top and right side, with start location in the bottom-left corner.}
	\label{fig:planar_nav_dems}
	\vspace{-0.5cm}
\end{figure}

\noindent\textbf{Experiment I: Simulated Planar Navigation.} We begin by testing our framework on a simple planar navigation problem, where a holonomic robot must reach a goal location while avoiding obstacles. We assume that the start, goal, and obstacle locations are known for a given planning problem, but the obstacle \textit{geometries} (ex. size, shape) are unknown. We want to learn an implicit distribution that captures the collision-free trajectories which lead to a particular goal. Here, we investigate two possible sources of empirical data: (1) sparsely populated point-distributions in free-space and (2) a set of expert trajectory distributions. The former can be seen as a stand-in for free-space measurements taken from a depth sensor, e.g., a lidar. In practice, we generate these points by sampling uniformly throughout the 2-D plane and rejecting points in collision. The expert trajectories are generated by running the \gls{sgpmp} planner at a high covariance for many (e.g., 500) iterations, producing a sparse set of trajectories for different goal locations and obstacle placements. We collect data for 512 different environments and produce trajectories from 15 random goal locations per environment, with 5 trajectories per goal. Both sources are depicted in ~\cref{fig:planar_nav_dems}. 
Here, the learned EBM can be expressed as $E_{\vtheta}(\vx, \{\vx^i_{obs}\}_{i=1}^N)$, where $\vx$ is a particular 2D-state, and $\vx_{obs}$ the position of an obstacle (here, $N=3$). 

\begin{figure}[!t]
	\centering
	\begin{minipage}{.49\columnwidth}
		\centering
		\includegraphics[width=\textwidth]{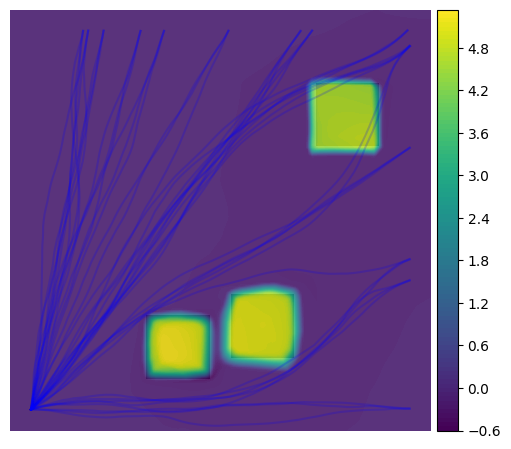}
	\end{minipage}
	\begin{minipage}{.49\columnwidth}
		\centering
		\includegraphics[width=\textwidth]{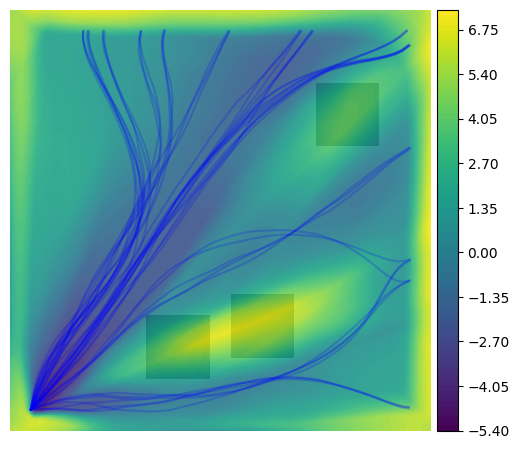}
	\end{minipage}	
    \caption{Learned obstacle-EBMs conditioned on novel obstacle locations. (Left) free-space point sampling and (right) expert trajectory distributions, with multi-goal planning solutions depicted by blue trajectories. Discontinuities and implicit obstacle surfaces are well captured using sparse free-space point-samples during training, whereas distributions of trajectory-based demonstrations can be captured neatly by the EBMs. The latter provides a convenient ``guiding" energy function for a new context, improving samples derived from multi-modal stochastic trajectory optimization.}
	\label{fig:planar_nav_ebm}
\end{figure}

\begin{figure}[!t]
	\centering
	\begin{minipage}{0.4\columnwidth}
		\centering
		\includegraphics[width=\textwidth]{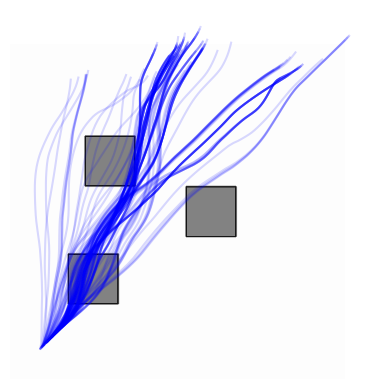}
	\end{minipage}
	\vspace{-0.3cm}
	\begin{minipage}{0.4\columnwidth}
		\centering
		\includegraphics[width=\textwidth]{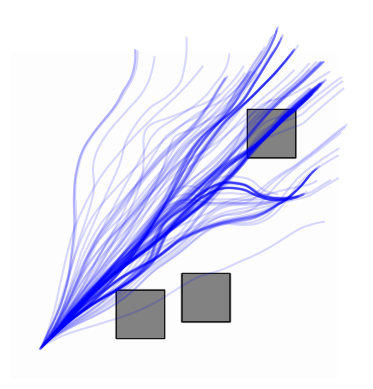}
	\end{minipage}
    \caption{Examples of trajectory traces using the behavioral cloning baseline on the planar navigation problem. Although rolled-out sample trajectories are generally well-behaved, they fail to avoid obstacles.}
	\label{fig:planar_nav_bc}
	\vspace{-0.55cm}
\end{figure}

Examples of the learned EBM for both cases are shown in \cref{fig:planar_nav_ebm}. The resulting energy functions, in either case, manage to effectively capture the demonstration distributions, conditioned on new obstacle locations. With \cref{sec:ebm_mo} in mind, we can generate trajectories near the modes of the full target distribution, accounting for smoothness and distance-to-goal, by first sampling from a structured prior and optimizing on the EBM to capture key modes of the distribution. We compare this method of implicit trajectory generation to a standard Behavioral Cloning (BC) baseline, where the learned policy outputs the current velocity, $\dot{\vq} = \vf(\vq; \mathcal{X}_o, \vx_g)$ which is conditioned on the set of obstacle poses $\mathcal{X}_o=\{\vx_o\}$ and the goal location $\vx_g$. The trajectories generated from this baseline can be seen in \cref{fig:planar_nav_bc}, obstacle avoidance is minimal. We perform a quantitative analysis on the EBM methods by measuring the success rate on the validation set as a function of optimization iterations needed by the planner (with the BC baseline fixed). This is shown in  Table \ref{tab:planar_nav}.\\

\resizebox{0.95\columnwidth}{!}{
    \centering
	\begin{tabular}{|c|c|c|c|c|c|}
	    \hline
		 Opt. iters. & 0 & 5 & 10 & 25 & 50 \\
		 \hline
		 EBM-Free-space & $0.556$ & $0.470$ & $0.643$ & $0.747$ & $0.852$ \\
		 \hline
		 EBM-Expert Traj. & $0.556$ & $0.690$ & $0.791$ & $0.847$ & $0.877$ \\
		 \hline
		 Behavioral cloning & $0.04$ & -- & -- & -- & -- \\
		 \hline
	\end{tabular}}
	\captionof{table}{Planar navigation: Average success rate per environment, as a function of optimization iterations. A planning trajectory is deemed successful if it ends within a radius of 1.5 from the goal, without hitting the underlying obstacles.}
	\label{tab:planar_nav}




\noindent\textbf{Experiment II: Simulated Planar Manipulator.}
This experiment studies the performance improvement in robotics manipulation tasks when introducing our expressive implicit priors in trajectory optimization. First, this experiment demonstrates how well the EBMs provide an informative prior landscape that biases the trajectory optimizer towards better solution regions than standard Gaussian priors. Second, it shows how the learned implicit priors on the task space result in superior performance when warm-starting the trajectory optimization, compared against explicit priors in configuration space, commonly done in literature \cite{koert2016demonstration}. 

\begin{figure}[t!]
	\centering
	\begin{minipage}{.49\columnwidth}
		\centering
		\includegraphics[width=.85\textwidth]{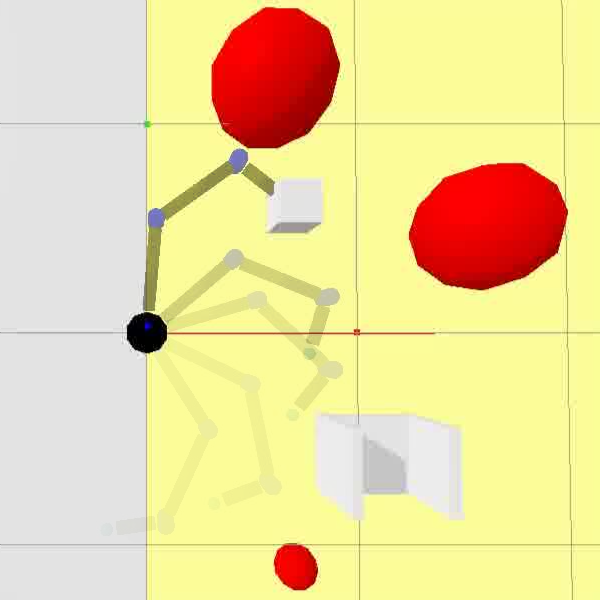}
	\end{minipage}
	\begin{minipage}{.49\columnwidth}
		\centering
		\includegraphics[width=.85\textwidth]{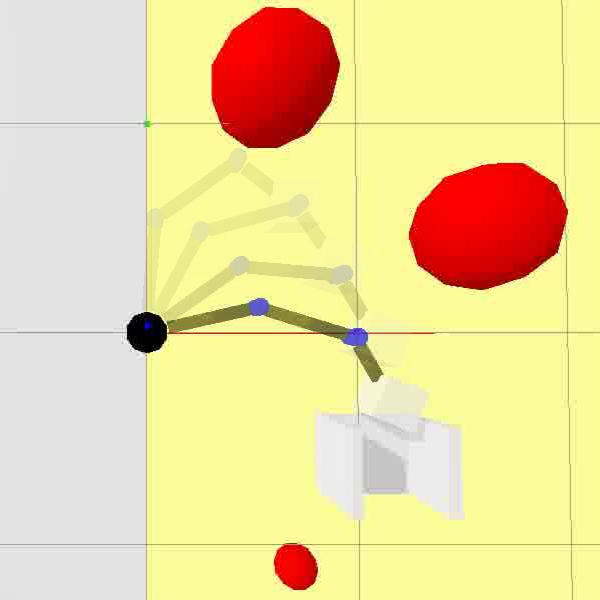}
	\end{minipage}
    \caption{Simulated planar arm executing the grasp \& insert tasks.}
	\label{fig:planar_se2bot}
	    \centering
    \resizebox{0.4999\textwidth}{!}{
	\begin{tabular}{|c|c|} 
	    \hline
		 Cost terms & Mathematical expression \\
		 \hline
		 Joint limit & $\max (\vq - \vq_{max}, 0) + \max (\vq_{min} - \vq, 0)$ \\
		 \hline
		 Sphere Obstacle Avoidance & $\sum_{o=1}^K \left\lVert\max (\vx_o + \epsilon - \vx(\vq), 0)\right\rVert^2$, $K$ is object number\\
		\hline
		Pose potential & $\frac{1}{2}\left\lVert \vx(\vq) - \vx_p \right\rVert^2$  \\
		\hline
	\end{tabular}}
	\captionof{table}{Elementary cost terms}
	\label{tab:costs}
\end{figure}

\begin{figure}
	\centering
	\begin{minipage}{.49\columnwidth}
		\centering
		\includegraphics[width=\textwidth]{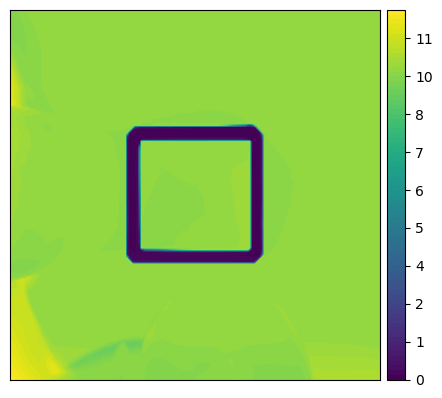}
	\end{minipage}
	\begin{minipage}{.49\columnwidth}
		\centering
		\includegraphics[width=\textwidth]{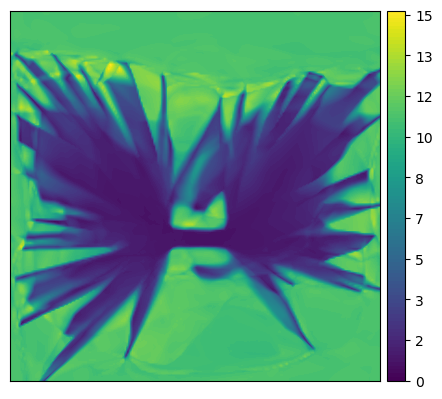}
	\end{minipage}
    \caption{The energy landscape of grasp \& insert learned models. (Left) Learned energy for grasping points.  (Right) Learned energy for placing in the cubby. Both energies are learned in the object-centric frame of the graspable object and the placing cubby. We generate the training data by motion planning. To cover all the possible placing trajectories, we run the motion planner given several initial end-effector poses.}
    
    
	\label{fig:pick_place_energy}

    \centering
    \includegraphics[width=\columnwidth]{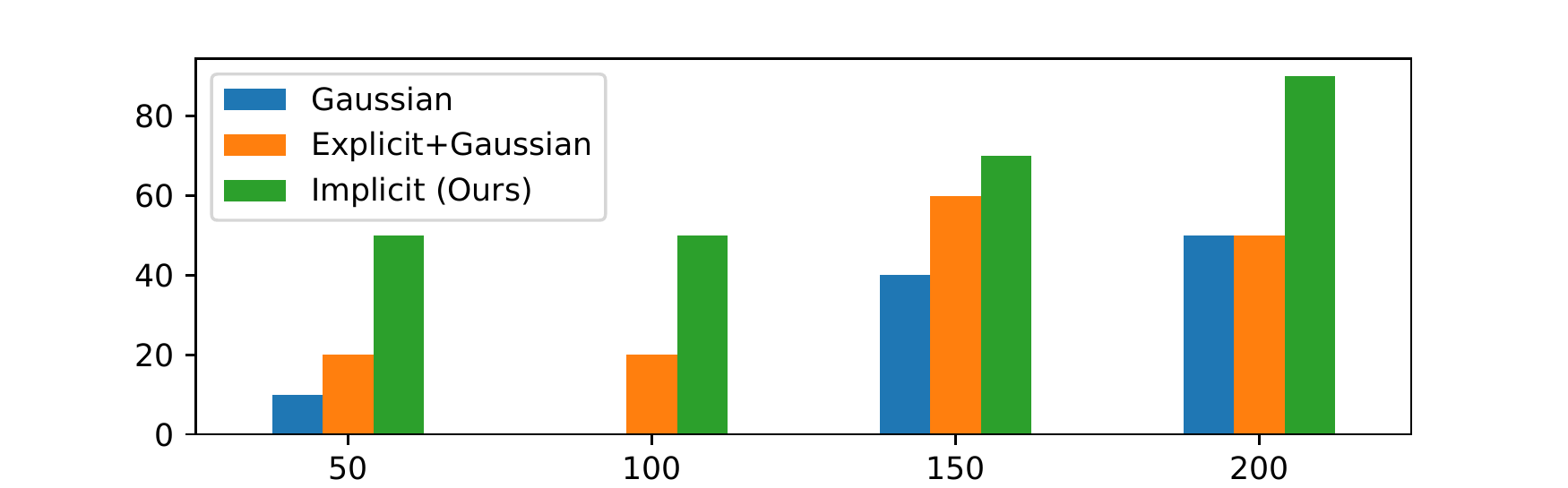}
    \caption{Success rate (\%) of grasp \& insert task over optimization steps. Surprisingly, even with few optimization steps of 50 \& 100, our implicit priors' success rate is already higher comparing to the baselines. This shows that using informative and expressive \gls{ebm}s can greatly assist the planner even with only few optimization steps. As optimization steps increase, our informative implicit priors still consistently outperforms previous methods.}
    \label{fig:planar_success_rate}
    \vspace{-15.pt}
\end{figure}

We create a simulated planar robot-arm with 3 dofs operated on a plane containing fixed red obstacles. The planner has to find an optimal low-jerk trajectory that guides the planar arm to grasp the cubic object and inserts the object into a 2-walled cubby without colliding with the cubby or the red obstacles (\cref{fig:planar_se2bot}). Our learned \gls{ebm}s represent the distribution of possible grasps and insertions that are used to generate trajectories that will bias the trajectory optimization problem towards the optimal solution. The learned energy distribution landscape is displayed in \cref{fig:pick_place_energy}, where the centers are the projected 2D-origin of the object-centric frame. We train both \gls{ebm}s using standard training methods as described in \cref{sec:learn_ebm}. When combining both implicit priors, the intuition is that the optimizer will plan the configuration trajectory that biases the end-effector toward the low-energy regions associated jointly with the most natural grasp and insert postures, given the current environmental setting. In all experiments, the grasp and insertion objectives are super-positioned with other standard costs such as smoothness, obstacle avoidance, and joint limits to completely define the trajectory optimization problem as in \eqref{eq:traj_opt}. 

We compare our method against the two baselines, where the first employs
Gaussian distributions as priors to define the object's grasp and insertion potentials in the task space (see ~\cref{tab:costs}) as commonly done in~\cite{ratliff2009chomp, kalakrishnan2011stomp}.
For the second baseline, we first learn a behavioral cloning model $\dot{\vq} = \vf(\vq; \vx_o, \vx_g)$ where $\vx_o, \vx_g$ represents the object's grasp pose and cubby pose, while the demonstrated configuration-space trajectories are generated for simplicity from a planner. Next, we warm start the optimizer with the trajectories sampled from the behavioral cloning model, where each trajectory is rolled out by integrating the model velocity output from the initial robot configuration. We expect that the second baseline is a stronger baseline due to the integration of an explicit prior.
Note that for the two baselines, we randomly sample a desired grasp pose on the object's surface for setting up the grasp potential cost, and we keep fixed the target placement pose  at the center of the cubby.

To show the strength of our implicit priors in trajectory optimization, we measure the success rate of successful grasping and insertions without collisions with obstacles or the target cubby. The success rates are reported over an increasing number of optimization steps in \cref{fig:planar_success_rate} to emphasize the benefit of our implicit priors in the multiple stages of the optimization, from a few steps till convergence. We run experiments with each method four times. Due to the stochasticity of the planner, we run 10 planning instances for each experiment and method, where at each instance we randomly sample a pair of a reachable cubic object pose and a target pose on the cubby. For all instances, the trajectory is fixed to 80 time-steps and the robot arm always starts from the same initial configuration.

We notice that most fail cases of the baselines are usually due to the collision with the cubby wall. It is because radial priors such as Gaussians do not encode the insertion dynamics, and the planner has to balance between the collision avoidance and the target pose objectives in a narrow placement target. We also observe that while warm starting with explicit priors does help the planner find a better solution than the Gaussian prior, sometimes the rollout trajectory of the behavioral cloning model suffers from covariate shift, that worsens the performance of the planner leading to fail cases.


    
\begin{figure}[t]
    \centering
	\begin{minipage}{.15\textwidth}
		\centering
		\includegraphics[width=.99\textwidth]{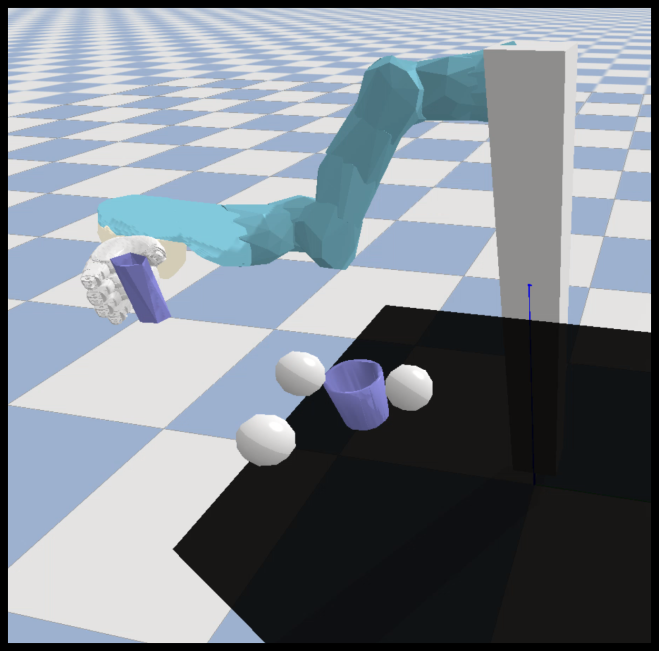}
	\end{minipage}
	\begin{minipage}{.15\textwidth}
		\centering
		\includegraphics[width=.99\textwidth]{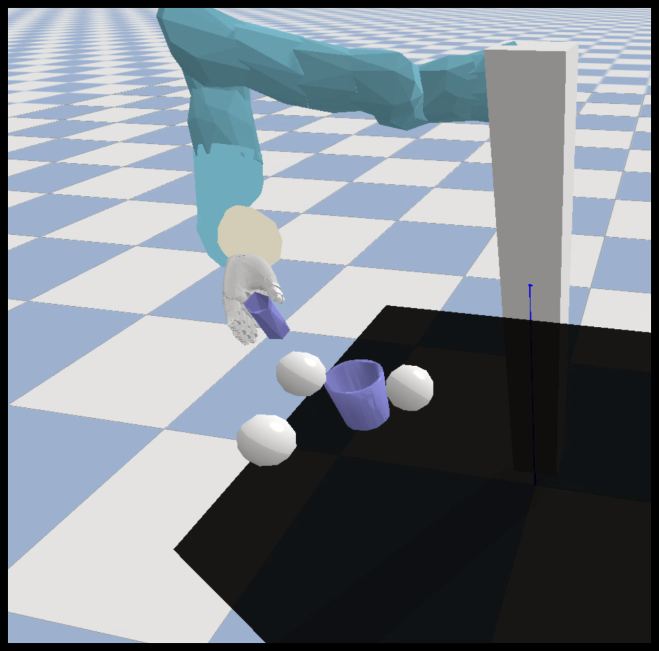}
	\end{minipage}
	\begin{minipage}{.15\textwidth}
		\centering
		\includegraphics[width=.99\textwidth]{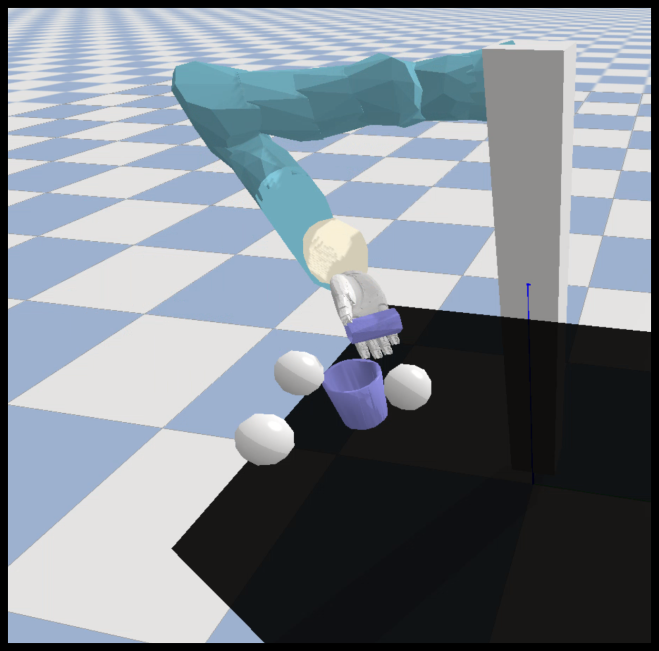}
	\end{minipage}
	\caption{A visualization of the pouring in cluttered task. The robot should avoid the obstacles, pour in the pot over the table and come back to the initial position.}
	\label{fig:pouring_task}
    \centering
    \includegraphics[width=.49\textwidth]{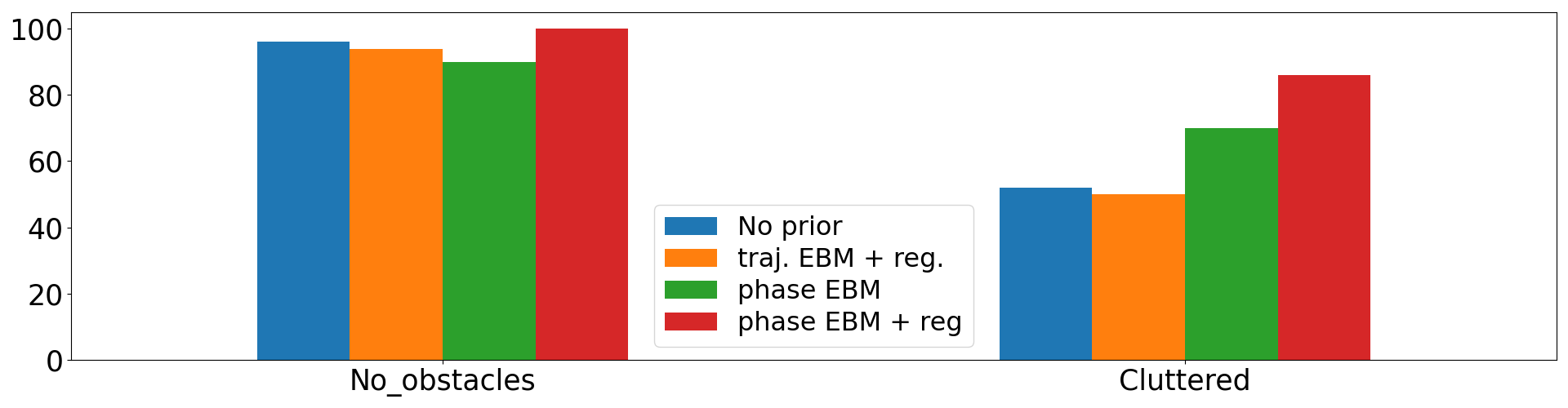}
    \caption{Success rate (\%) for the pouring task. (Left): Experiment without collision obstacles. The success rate of the motion optimization without any learned \gls{ebm} is almost the maximum. We observe that the phase \gls{ebm} with regularization can improve the performance slightly. (Right): Experiment with obstacles in the environment. We observe a clear benefit of using phase-based \gls{ebm} in contrast with trajectory-based \gls{ebm}. Training the \gls{ebm} in high dimensional space requires too many samples, and it is difficult to get smooth \gls{ebm}s representing the demonstrations. The \gls{ebm} trained with regularization improves the obtained results with respect to non-regularized \gls{ebm}. For any initial position of the trajectory particles, the regularized \gls{ebm} provides an informative gradient towards the demonstrations, while in the case of non-regularized \gls{ebm}, we found the gradient was informative only in the close vicinity of the demonstrations.}
    \label{fig:pouring_results}
    \vspace{-20pt}
\end{figure}

\noindent\textbf{Experiment III: Robot pouring amid obstacles.} This experiment studies the performance gains of introducing learned implicit priors for a more complex manipulation task in a higher-dimensional space. Namely, we investigate the integration of \gls{ebm}s in trajectory optimization for a pouring task in the presence of obstacles with a 7dof LWR-Kuka robot arm. This experiment investigates (i) the benefit of including smoothness regularization in the \gls{ebm} training, (ii) the advantages of phase-conditioned \gls{ebm} w.r.t. learning the \gls{ebm} in trajectory space, and (iii) the generalization of our \gls{ebm}s in the context of the pouring task regarding arbitrary pouring places, and in the presence of obstacles. Instances of our method's performance are available in \cref{fig:pouring_task} for the simulated task, and \cref{fig:main_fig} for our zero-transfer to the actual robotic setup. We add additional details on the problem setup in the Appendix on the \href{https://sites.google.com/view/implicit-priors}{project site}. 

To learn the pouring \gls{ebm}, we recorded $500$ trajectory demonstrations of the pouring task. The demonstrations were generated using a set of handcrafted policies. The demonstrations were initialized in arbitrary initial configurations and performed the pouring from different positions, generating a multimodal distribution of demonstrated trajectories. The demonstrations have been directly recorded for the glass pose and centered in the frame of the pouring pot. To properly encode the temporal information in the data, we learn a time conditioned \gls{ebm}, $E_{\vtheta}(\vx|\alpha)$. In our problem, $\vx$ is a 6-dimensional state, representing the 3D position of the bottom and tip of the glass w.r.t. the pouring pot frame. Centering the \gls{ebm} to the pouring pot's frame allows us to generalize the \gls{ebm} to arbitrary pot poses. Additionally, we include the denoising regularization and compare its performance and compare to a baseline without the proposed regularization.

We compare against three baselines. First, a solver without any prior, to appropriately evaluate the benefits of adding guiding priors. Second,  a phase-conditioned \gls{ebm} without smoothness regularization combined with the optimizer. Third, an \gls{ebm} that is directly learned in the trajectory space to investigate the benefit of phase-conditioned \gls{ebm}s in trajectory optimization. Given the learned pouring prior, we optimize the task trajectory by gradient descent. The objective function is defined by the composition of a set of cost functions--fixed initial configuration, fixed target configuration, trajectory smoothness, obstacle avoidance, keep the glass pointing up to avoid spilling and pour inside the pot~(\cref{tab:costs}). The learned \gls{ebm} is added as an additional factor in the optimization problem. We optimize using a tempering scheme, giving more importance to the prior at the beginning and reducing its influence at the end of the optimization process.


To evaluate the pouring \gls{ebm}, we measure the success rate of the motion optimization to perform the pouring task successfully. We consider an episode to be successful if it satisfies three conditions: (i) there are no collisions, (ii) the robot does not spill outside the pot, and (iii) the robot pours inside the pot. We report performance both in obstructed and obstacle-free environments. We run $50$ episodes for each case. we randomize the position of the pouring pot and the obstacles on each episode. The obtained results are presented in \cref{fig:pouring_results}. We can clearly observe the benefit of both integrating smoothing regularizers in the training of the \gls{ebm} and modelling the \gls{ebm}s as phase-conditioned models. The presented results are complemented with the attached video for the simulated tasks, which additionally showcases our successful transfer to the real robotic system. Note that our learned priors do not suffer for sim2real mismatch, as the only requirement is the knowledge of the object's pose, which in our case, we resolve through motion tracking. 



\section{Discussion and Conclusions}
Energy-based models (EBM) have multiple desirable properties that make them obvious candidates to be integrated as data-driven priors in robotics. \gls{ebm}s  can represent highly multimodal distributions, and due to their implicit nature, they can naturally be composed and integrated into trajectory optimization problems. Nevertheless, the high dimensionality and variability of the tasks in robotics hinder the learning of generalizable \gls{ebm}s that could apply to a wide range of complex manipulation tasks. In this paper, we demonstrate that learning \gls{ebm}s as implicit priors can guide motion optimization for common robotics problems. We introduce a set of modeling choices to represent generalizable \gls{ebm}s that could be adapted to an arbitrary set of manipulation tasks. We propose alternative training losses to learn smoother \gls{ebm}s that could provide well-conditioned gradients into optimization problems. Moreover, we introduce novel structured sampling procedures to combine data-driven models with explicit priors. Our experimental results validated the efficacy and applicability of our proposed algorithmic choices for motion optimization. In future works, we aim to explore the problem of learning conditioned \gls{ebm}s given higher dimensional contexts, such as images or text.

\bibliographystyle{ieeetr}
\bibliography{bib_EBM,bib_MP, bib_learning4MP}

\onecolumn
\setlength{\parindent}{10pt}

\begin{appendices}

\section{Stochastic Gaussian Process Motion Planning}
\label{app:stoch_gpmp}

\subsection{Gaussian Process Trajectory Prior}
\label{sec:gp_prior}
We can define a prior on the space of smooth, continuous-time trajectories $\stateb(t)$ using a Gaussian Process~\cite{barfoot2014batch, mukadam2018continuous}:  $\stateb(t) \sim \mathcal{GP}(\mub(t),\, \Kb(t,t'))$, with mean function $\mub$ and covariance function $\Kb$. This distribution is parameterized by a discrete set of support states $\vtau = [\stateb_0, \stateb_1, ..., \stateb_N]^\top$, which constitutes a discrete-time trajectory defined along times $\tb=[t_0, t_1, ..., t_N]$. We can then consider the distribution over $\vtau$ as $p(\vtau) = \mathcal{N}(\mub, \Kb)$, with mean $\mub =[\mub(t_0), ..., \mub(t_N)]^\top$ and covariance matrix $\Kb = [\Kb(t_n,t_m)]\big|_{nm, 0\leq n,m \leq N}$. \vspace{5pt}

As described in \cite{barfoot2014batch, sarkka2013spatiotemporal}, a GP prior can be constructed in a principled manner by considering the structured covariance function created by a linear time-varying stochastic differential equation (LTV-SDE)
\begin{align}\label{eq:gp_dyn}
    \dot{\stateb} = \textbf{A}(t)\stateb(t) + \textbf{u}(t) + \textbf{F}(t)\textbf{w}(t)
\end{align}
where $\textbf{u}(t)$ is the control input, $\textbf{A}(t)$ and $\textbf{F}(t)$ the time-varying system matrices, and $\textbf{w}(t)$ is a disturbance following the white-noise process $\textbf{w}(t) \sim \mathcal{GP}(0, \textbf{Q}_c \delta(t-t'))$, where $\textbf{Q}_c$ is the power-spectral density matrix.

\begin{figure}
    \centering
    \includegraphics[width=0.7\textwidth]{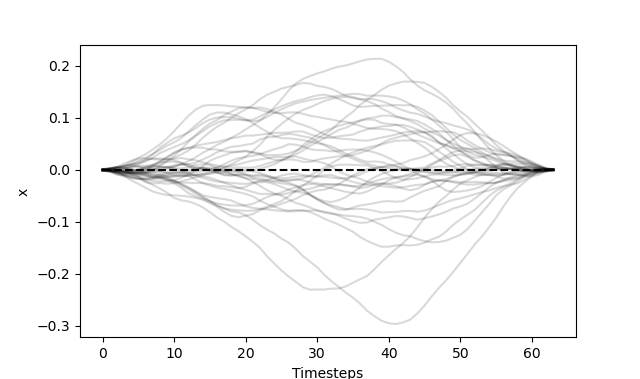}
    \caption{An example GP prior for 1D trajectories. The dashed line is
the mean constant trajectory. The 32 solid lines are sampled trajectories from
the GP prior. }
    \label{fig:gp_samples}
\end{figure}

Given a prior on start state $p_s(\vx)=\mathcal{N}(\vmu_s, \vSigma_s)$ and end-goal position  $p_g(\vx)=\mathcal{N}(\vmu_g, \vSigma_g)$, the trajectory prior can be defined as the product of unary and binary factors

\begin{align}\label{eq:gp_prior}
    p_{\textrm{GP}}(\vtau) \, &\propto \,  \exp\big(-\frac{1}{2}\big|\big| \vtau - \vmu\big|\big|^2_{\Kb}\big) \\
    &\propto \, p_s(\vx_0)\  p_g(\vx_N)  \prod_{i=0}^{N-1} p_i^{gp} (\vx_i, \vx_{i+1})
\end{align}
Following \cite{barfoot2014batch}, each factor is then defined as
\begin{align}
\begin{split}
    p^{gp} (\bm\stateb_i, \bm\stateb_{i+1}) &\propto \exp \Big( \hspace{-1mm}-\frac{1}{2} \| \mathbf{\Phi}_{i,i+1} (\bm\stateb_i - \vmu_i) - (\bm\stateb_{i+1}  - \vmu_{i + 1})  \|^2_{\mathbf{Q}_{i,i+1}} \Big), \\
    p_s(\vx_0) &\propto \exp \big( \hspace{-1mm}-\frac{1}{2}  \| \vx_0 - \vmu_s\|^2_{\vSigma_s}  \big), \\  
    p_g(\vx_N) &\propto \exp \big( \hspace{-1mm}-\frac{1}{2}  \|\vx_N - \vmu_g\|^2_{\vSigma_g}  \big)
\end{split}
\end{align}
where $\mathbf{\Phi}_{i,i+1} $ is the state transition matrix and $\mathbf{Q}_{i,i+1}$ the variance between times $t_i$ and $t_{i+1}$. For a first-order LTV-SDE, where the state consists of position and velocity terms $\vx = [\textbf{x}, \dot{\textbf{x}}]^\top$, $\textbf{x}, \dot{\textbf{x}} \in \mathbb{R}^d$, the state transition and covariance matrices equate to

\begin{gather*}
\small{
\mathbf{\Phi}_{i,i+1} = \begin{bmatrix}
\mathbf{I} & \Delta t_i\mathbf{I} \\ 
\mathbf{0} & \mathbf{I}
\end{bmatrix}, \quad
\mathbf{Q}_{i,i+1} = \begin{bmatrix}
\frac{1}{3} \Delta t_i^3 \mathbf{Q}_c &
\frac{1}{2} \Delta t_i^2 \mathbf{Q}_c \\ 
\frac{1}{2} \Delta t_i^2 \mathbf{Q}_c &
\Delta t_i \mathbf{Q}_c
\end{bmatrix}
}
\end{gather*}
where $\Delta t_i = t_{i+1} - t_i$. Sampling from this distribution using the time-correlated GP covariance results in trajectories which have low variance near the start and goal positions (depending on the values of $\vSigma_s$ and $\vSigma_g$), and higher variance near the middle of the horizon length. An example of smooth, goal-directed trajectories sampled from a one-dimensional GP-prior is depicted in \cref{fig:gp_samples}. 

\subsection{Stochastic Optimization for Motion Planning}
\label{sec:mp_derivation}

As described in the main text, we can express the motion-planning problem as finding the trajectory which minimizes the negative log-likelihood of a factored probability distribution
\begin{align}
    \vtau^* &= \arg \min_{\vtau} - \log p_{\textrm{GP}}(\vtau) \prod_k p_k(\vtau)
\end{align}
with non-negative functions $p_k(\vtau) \propto \exp(-E_k(\vtau))$ encoding additional constraints, such as obstacle collision costs and joint limit violations.  

Alternatively, we can view the problem of \textit{stochastic} motion planning in terms of distribution matching, where we seek to match a proposal trajectory distribution $q(\vtau)$ to some target distribution $q^*(\vtau)$. This is perhaps most clearly described by the information theoretic derivation of model-predictive path integral (MPPI) control in \cite{williams2017information}. Here, we consider a system with general discrete-time dynamics defined by $X=F(\textbf{x}_0, V)$, for a state-sequence $X=(\textbf{x}_0, \textbf{x}_1, ..., \textbf{x}_N)$, control-trajectories $V=(\textbf{v}_0, \textbf{v}_1, ..., \textbf{v}_N)$, and start-state $\textbf{x}_0$. The target distribution over control-trajectories $V=(\textbf{v}_0, \textbf{v}_1, ..., \textbf{v}_N)$ is then maximum-entropy Gibbs distribution
\begin{align}
    q^*(V) = \frac{1}{\eta}\exp\left(-\frac{1}{\lambda}S(V)\right) p(V)
\end{align}
where $p(V) = \mathcal{N}(\textbf{0}, \vSigma)$, a zero-mean centered normal distribution, $S(\cdot)$ the cost function, $\lambda$ a scaling term, and $\eta$ a normalizing factor. Note that $q^*(V)$ can be interpreted as a \textit{posterior} distribution, with a likelihood function provided by $\exp\left(-\frac{1}{\lambda}S(V)\right)$ and prior $p(V)$.  Given the open-loop characterization of the system, we can consider distributions directly over $V$, leaving the state-dynamics $F(\cdot, \cdot)$ as part of the cost-evaluation function $S = E \circ F(\textbf{x}_0, V)$, for some function $E$. 

For a kinematic system, the control sequence can be defined as a time-derivative of state, ex. $V = (\dot{\textbf{x}}_0, \dot{\textbf{x}}_1, ..., \dot{\textbf{x}}_N)$, and we can instead consider the joint distribution over \textit{both} state-trajectories and their derivatives: 
$\vtau = (X, V) = (\textbf{x}_0, \dot{\textbf{x}}_0, \textbf{x}_1, \dot{\textbf{x}}_1, ..., \textbf{x}_N, \dot{\textbf{x}}_N)$. 
As such, we can similarly define the optimal distribution to be
\begin{align}
    q^*(\vtau) = \frac{1}{\eta}\exp\left(-\frac{1}{\lambda}E(\vtau)\right) p_F(\vtau)
\end{align}
where the dynamics $F$ are now integrated into the prior distribution $p_F(\vtau)$, describing the un-controlled stochastic dynamics. Fortunately, we have already derived an example of such a distribution in \cref{eq:gp_prior}, using the dynamics in \cref{eq:gp_dyn}. Leveraging the time-correlated prior covariance $\Kb$, we can define the nominal dynamics distribution as $p_F(\vtau) = \mathcal{N}(\textbf{0}, \Kb)$.

Using the proposal distribution $q(\vtau; \thetab) = \mathcal{N}(\vmu, \Kb)$ with parameters $\thetab = (\vmu, \Kb)$, we can then minimize the KL-divergence with the target distribution
\begin{align}
    \thetab^* &= \arg \min_{\thetab} \kldiv{q^*(\vtau)}{q(\vtau; \thetab)}
\end{align}

Introducing the nominal trajectory distribution $p_F(\vtau)$ into the equation
\begin{align}\label{eq:objective}
    \thetab^* &= \arg \min_{\thetab} \int q^*(\vtau) \log \frac{q^*(\vtau)}{q(\vtau; \thetab)} \frac{p_F(\vtau)}{p_F(\vtau)} d\vtau \\
    &= \arg \min_{\thetab} - \int q^*(\vtau) \log \frac{q(\vtau; \thetab)}{p_F(\vtau)} d\vtau
\end{align}
where the cross-entropy objective in the last line results from independence of the entropy $\int q^*(\vtau) \log \frac{q^*(\vtau)}{p_F(\vtau)} d\vtau$ from $\thetab$.

In order to derive a practical algorithm for solving this objective, we can simplify the problem by optimizing for only the mean trajectory $\thetab = \vmu$, keeping $\Kb$ as a fixed covariance. Sampling trajectories from $q^*(\vtau)$ directly is typically intractable. Following the derivation in \cite{williams2017information}, we can instead apply importance sampling such that the proposal distribution $q(\vtau; \thetab)$ can be used for evaluating the expectation in \cref{eq:objective}. We then arrive at the update rule to the proposal distribution parameter
\begin{align}
    \vmu_{new} = \vmu + \frac{\expect{\vtau \sim q(\vtau; \vmu)}{\exp \left(-\frac{1}{\lambda} E(\vtau) + \alpha(\vtau) \right) (\vtau - \vmu)}}{\expect{\vtau \sim q(\vtau; \vmu)}{\exp \left(-\frac{1}{\lambda} E(\vtau) + \alpha(\vtau)\right)}}
\end{align}

where the term $\alpha(\vtau) = -\frac{1}{2} \vmu \Kb^{-1} \vtau$ results from the importance-sampling factor. 

The expectations are approximated via Monte Carlo sampling, where $K$ trajectories are drawn directly from the proposal distribution:  $\{\vtau^k\}_{k=0}^K \sim q(\vtau; \vmu)$. Using the notation $\Tilde{E}(\vtau) = E(\vtau) - \lambda \alpha(\vtau)$, we have
\begin{align}
        \vmu_{new} &= \vmu + \frac{\sum_{k=0}^K \exp \left(-\frac{1}{\lambda} \Tilde{E}(\vtau^k) \right) (\vtau^k - \vmu)}{\sum_{k=0}^K \exp \left(-\frac{1}{\lambda} \Tilde{E}(\vtau^k)\right)} \\
        &= \vmu + \sum_{k=0}^K w^k (\vtau^k - \vmu)
\end{align}
where $w^k$ is the softmax-weight for $\vtau$ with respect to the other samples. Similarly to \cite{wagener2019online}, we can also incorporate a step-size, or smoothing parameter $\gamma\in[0,1]$
\begin{align}
    \vmu_{new} &= \vmu + \gamma \sum_{k=0}^K w^k (\vtau^k - \vmu)\\
    &= (1-\gamma)\vmu + \gamma \sum_{k=0}^K w^k \vtau^k.
\end{align}
This update can be then applied in an iterative fashion until a converged trajectory distribution is reached. The resulting \textit{posterior} distribution approximation, $q_{final}(\vtau)$, will be parameterized by a mean trajectory $\mub_{final} = [\stateb_0, \stateb_1, ..., \stateb_N]^\top$ (where a state $\stateb_i$ includes position and velocity elements, $\stateb_i = (\textbf{x}_i, \dot{\textbf{x}}_i)$, for example). This trajectory provides the motion plan in configuration-space to be used for execution on the system.

Note that, unlike MPPI, there is no need to perform \textit{roll-outs} of a dynamics model to generate trajectory samples. Since the goal-directed stochastic dynamics are integrated into the proposal distribution via the time-correlated covariance matrix $\Kb$, we can sample trajectories directly from the system to acquire sequences of positions, velocities, etc. This avoids the linear-with-time complexity of sequential Markov sampling using single-step dynamics.  Furthermore, the sampling-based optimization scheme of StochGPMP does not require that the cost function $E(\cdot)$ to be differentiable. For motion planning problems, this permits the use of hard constraints such as discontinuous occupancy functions (to be used as obstacle costs) or joint limits, for instance. 

\subsection{GP Interpolation}

A main advantage of maintaining a Gaussian Process trajectory distribution is the ability to query the posterior at \textit{any time} within the planning horizon, \ie\  $t_0 < t' < t_N$. As mentioned in \cite{dong2016motion}, this can be used to densify the mean trajectory $\mub$ of the optimized proposal distribution, smoothly interpolating motion commands between timesteps. Specifically, given an optimized mean trajectory $\mub_{final}$ over times $\tb=[t_0, t_1, ..., t_N]$, we can interpolate to a dense time series $\tb_d=[t_0, t'_1, ..., t'_{M-1}, t'_M]$ where $N < M$ and $t'_{j+1}-t'_{j} < t_{i+1}-t_{i}\ \forall \ i, j$. This is done by conditioning the mean of the posterior~\cite{rasmussen2003gaussian}
\begin{align}
    \vmu_{d} = \Kb(\tb_{d},\ \tb)\,\Kb^{-1} \vmu ,
\end{align}
where the kernel matrix $\Kb(\tb_{d},\ \tb) = [\Kb(t'_m,t_n)]\big|_{0\leq m \leq M, 0\leq n \leq N}$ can be constructed as in \cref{sec:gp_prior} by considering the Markovian dependencies between neighboring states for transition matrices $\mathbf{\Phi}_{i,j}$, $\mathbf{\Phi}_{j,i+1}$  and covariances $\mathbf{Q}_{i,j}$, $\mathbf{Q}_{j,i+1}$, where $t_i \leq t'_j \leq t_{i+1}$. 
The interpolation step may be desirable for execution on systems requiring higher temporal frequencies. Also, the projection term $\Kb(\tb_{d},\ \tb)\,\Kb^{-1}$ can be pre-computed and stored prior to planning. As with $\Kb$, this matrix is sparse and can benefit from efficient storage and matrix-multiplication using libraries for sparse linear algebra. 

In GPMP~\cite{dong2016motion,mukadam2018continuous}, interpolation is applied in-the-loop to finely resolve collision costs, before projecting the evaluations to the support states. This may warrant a similar approach in StochGPMP, where densification of the mean trajectory $\mub$ would allow sampling of more finely resolved trajectories for evaluation (note that the softmax weights $w^k$ evaluate an entire trajectory $\vtau^k$, so projection back to support states would be unnecessary here). However, since we would still need to sample dense trajectories from the full distribution for evaluation, this approach would not provide any significant gain in space or runtime complexity. Therefore, the temporal resolution of the optimized mean trajectory $\mub$ is initially chosen to adequately resolve obstacle features. 

\subsection{Algorithm}
\label{sec:alg}

Given a start state $\stateb_0\in\mathbb{R}^b$ and goal state $\stateb_g\in\mathbb{R}^b$ defined in configuration space having $b$ dimensions, as well as a cost-function $E(\cdot)$, we seek to find the optimal trajectory $\mub^*$ which avoids high-cost regions and defines a motion plan which reaches the goal.

As described in the main body of the paper, the function $E(\cdot)$ can be composed as the sum of different functions, defining separate constraints and penalties \ie\ $E(\cdot) = \sum_p E_p(\cdot)$. For encoding task-based constraints, these cost components typically consist of a function $G:\mathbb{R}^{\ell\times N} \rightarrow \mathbb{R}$, where $\ell$ is the dimension of the task space and $N$ the trajectory length, as well as the forward-kinematic mapping $FK: \mathbb{R}^{b\times N} \rightarrow \mathbb{R}^{\ell \times N}$, such that the cost component function is the composition $E_p(\cdot) = G(FK(\cdot))$.  It should be emphasized that neither the function $G$ or $FK$ need to be differentiable for sampling-based optimization. It is often desirable to define a goal position within task space, such as the desired end-effector pose $\mathbf{p}_g\in SE(3)$ and velocity $\dot{\mathbf{p}}_g\in SE(3)$ at the goal for a reaching task on a robot manipulator. In this, case one may simply use inverse kinematics to initialize the goal state in configuration space: $\stateb_g = IK(\mathbf{p}_g, \dot{\mathbf{p}}_g)$.
 
We additionally need to define covariance matrices $\vSigma_s\in \mathbb{R}^{b\times b}$ and $\vSigma_g\in \mathbb{R}^{b\times b}$ for the start and goal states, respectively. These may explicitly represent the actual uncertainty in these positions, or they can be viewed as inverted-weights which scale the importance of generating samples which start at $\stateb_0$ and end at $\stateb_g$. To initialize the sampling distribution $q(\vtau; \mub, \Kb)$, we first set the mean $\mub$ to be the straightline trajectory between $\stateb_0$ and $\stateb_g$, equidistant along the time sequence $\tb=[t_0, t_1, ..., t_N]$ and with constant velocity. The GP covariance matrix is initialized at the start of the optimization using the procedure in \cref{sec:gp_prior}. Although $\Kb$ remains fixed in the current implementation, it is conceivable to gradually decrease the GP variance by scaling the $\mathbf{Q}_c$ matrix according to some schedule, thereby decreasing the entropy of the distribution for denser sampling and improved estimation of $q^*$. 
In \cref{algo:svtrajopt_unimode}, we describe Stochastic Gaussian-Process Motion Planning (StochGPMP), an algorithm for single-trajectory/uni-modal trajectory optimization.

{\SetAlgoNoLine%
  \begin{algorithm}[ht!]
    \DontPrintSemicolon 
    \KwIn{
        Start state $\stateb_0$,
        goal state $\stateb_g$,
        cost function $E(\cdot)$,
        temperature $\lambda$,
        start-factor cov. $\vSigma_0$,
        goal-factor cov. $\vSigma_g$,
        $\textbf{Q}_c$,
        step-size $\gamma$,
        time sequence $\tb$,
        dense time sequence $\tb_d$ (optional)
        }
    \vspace{10pt}
    \tcp{Initialize mean trajectory}
    $\vmu \leftarrow InitMean(\vx_0, \vx_g, \tb)$\\
    \tcp{Construct GP covariance}
    $\Kb \leftarrow InitCov(\vSigma_0, \vSigma_g, \textbf{Q}_c, \tb)$\\[5pt]
    \While{Not Converged} {
        Sample $\{\vtau^k\}_{k=1}^{K} \sim q(\vtau; \vmu, \Kb)$\\[5pt]
        \tcp{Batched computation}
        \ForPar{$i = 1, 2, ...,K$}{\vspace{2pt} 
            \tcp{Evaluate trajectory cost}
            $E^k = E(\vtau^k)$\\
        }
        \ForPar{$i = 1, 2, ...,K$}{\vspace{2pt}
            \tcp{Compute weights including important-sampling factor}
            $w^k \leftarrow \frac{\exp \left(-\frac{1}{\lambda} \Tilde{E}(\vtau^k)\right)}{\sum_{j=0}^K \exp \left(-\frac{1}{\lambda} \Tilde{E}(\vtau^j)\right)}$\\
        }
        \tcp{Update proposal distribution}
        $\vmu \leftarrow \vmu + \gamma \sum_{k=0}^K w^k (\vtau^k - \vmu)$\\[5pt]
    }\vspace{5pt}
    \tcp{Optional: GP-interpolation for trajectory densification} 
    $\vmu_{d} \leftarrow \Kb(\tb_{d},\ \tb)\,\Kb^{-1} \vmu$\\[5pt]
    \caption{StochGPMP}
    \label{algo:svtrajopt_unimode}
  \end{algorithm}}%

{\SetAlgoNoLine%
  \begin{algorithm}[h]
    \DontPrintSemicolon 
    \KwIn{
        Start state $\stateb_0$,
        goal states $\mathbf{X}_g$,
        cost function $E(\cdot)$,
        temperature $\lambda$,
        start cov. matrix $\vSigma_0$,
        goal cov. matrices $\bm{\Xi}_g$,
        planning GP variance matrix $\textbf{Q}_c$,
        prior GP variance matrices $\mathbf{Q}_{c,0}$,
        step-size $\gamma$,
        time sequence $\tb$,
        dense time sequence $\tb_d$ (optional)
        }
    \vspace{10pt}
    \tcp{Prior const-vel. mean trajectories}
    $\mutens_0 \leftarrow InitMeans(\vx_0, \mathbf{X}_g, \tb)$\\[5pt]
    \tcp{Prior GP cov. matrices}
    $\bm{K}_0 \leftarrow InitCovs(\vSigma_0, \bm{\Xi}_g, \mathbf{Q}_{c,0}, \tb)$\\[5pt]
    \tcp{Initialize mean trajectories}
    $\mutens \sim \mathcal{N}(\mutens_0, \bm{K}_0)$ \\[5pt]
    \tcp{Construct planning cov. matrices}
    $\bm{K} \leftarrow InitCovs(\vSigma_0, \bm{\Xi}_g, \textbf{Q}_c, \tb)$\\[5pt]
    \For{ $ iter = 1:iter_{max}$} {
        \vspace{5pt}
        \tcp{Sample trajectories}
        $\mathbf{T} \sim q(\mathbf{T}; \mutens, \bm{K})$\\[5pt]
        \tcp{Evaluate trajectory costs}
        $E_{\mathbf{T}} \leftarrow E(\mathbf{T})$\\[5pt]
        \tcp{Compute weights across sampling dimension}
        $\mathbf{W} \leftarrow softmax(-\frac{1}{\lambda} E_{\mathbf{T}})$ \\[5pt]
        \tcp{Update proposal distribution}
        $\mutens \leftarrow \mutens + \gamma \mathbf{W} \otimes (\mathbf{T} - \mutens)$\\[5pt]
    }\vspace{5pt}
    \tcp{Evaluate solutions}
    $E_{\mutens} \leftarrow E(\mutens)$\\[5pt]
    \tcp{Pick best trajectory}
    $i^* \leftarrow \argmin_i [E_i \in E_{\mutens}]$ \\[5pt]
    $\mub^* \leftarrow \mub_{i^*} \in \mutens$ \\[5pt]
    \tcp{Optional: GP-interpolation for trajectory densification} 
    $\vmu_{d} \leftarrow \Kb(\tb_{d},\ \tb)\,\Kb_*^{-1} \vmu^*$\\[5pt]
    \caption{Multi-StochGPMP}
    \label{algo:svtrajopt_multimode}
  \end{algorithm}}%

\subsection{Multi-goal, Multi-modal Motion Planning}

Motion planning is generally a non-convex optimization problem, where different homotopy classes can be induced by the presence of obstacles~\cite{kolur2019online, lambert2020stein}. Furthermore, in the case of multi-goal planning, we may have a distribution of possible goals for a given environment or set of obstacles. Thus, most motion planning approaches will only return an approximate locally optimal solution. We would ideally like to leverage parallel computation to efficiently provide many feasible solutions for the multi-goal and multi-modal setting. 

We can implement a vectorized version of \cref{algo:svtrajopt_unimode} for the multi-goal, multi-modal setting using PyTorch~\cite{paszke2019pytorch}. This permits GPU parallelization across different goals, motion plans, trajectory sample generation and evaluation. For a given environment, we may have $N_g$ number of goal states: $\mathbf{X}_g=[\stateb_g^1, \stateb_g^2, ..., \stateb_g^{N_g}]^\top \in \mathbb{R}^{N_g\times b}$, with goal covariance matrices $\bm{\Xi}_g = [\vSigma_g^1, \vSigma_g^2, ..., \vSigma_g^{N_g}]^\top \in \mathbb{R}^{N_g \times b \times b}$. For each goal state, we may wish to independently optimize for multiple motion plans of the same length, which can be represented by the the mean trajectory tensor $\mutens^{n_g} = [\mub_{n_g}^1, \mub_{n_g}^2, ..., \mub_{n_g}^{N_p}]^\top\in \mathbb{R}^{N_p \times N \times b}$, where $n_g\in [1, 2, ..., N_g]$ is the goal index, and $N_p$ the number of plans-per-goal. This can yet be further expanded to include plans across all goals: $\mutens = [\mutens^1, \mutens^2, ..., \mutens^{N_g}]^\top \in \mathbb{R}^{N_g \times N_p \times N \times b}$. In this case, the collection of planning trajectories $\mutens$ is the parameter to be optimized. We can also construct a global covariance matrix, which might contain independent GP-covariances for each goal location (with possibly different noise levels set by $\mathbf{Q}_c$): $\bm{K} = [\Kb^1, \Kb^2, ..., \Kb^{N_g}]^\top$. We can then define a vectorized proposal distribution $q$ from which we can draw trajectory samples across goals and motion plans: $\mathbf{T} \sim q(\mathbf{T}; \mutens, \bm{K})$ where $\mathbf{T} \in \mathbb{R}^{N_g \times N_p \times K \times N \times b}$ for $K$ number of samples per motion plan. These samples can then be evaluated by the cost-function $E(\cdot)$, provided that  this is also suited to vectorized processing (which is the case for a neural network model in PyTorch, for instance). 

To initialize the mean trajectories $\mutens$, we randomly sample from a starting prior distribution:  $\mutens \sim \mathcal{N}(\mutens_0, \bm{K}_0)$, where $\mutens_0 = [\mub_0^1, \mub_0^2, ..., \mub_0^{N_g}]^\top \in \mathbb{R}^{N_g \times N \times b}$ are the constant-velocity, straight-line trajectories from the start state to each goal state, and $\bm{K}_0 = [\Kb_0^1, \Kb_0^2, ..., \Kb_0^{N_g}]^\top \in \mathbb{R}^{N_g \times (N \times b) \times (N \times b)}$ the independent covariance matrices. These in turn can be constructed from goal-specific covariances $\bm{\Xi}_g$ and state-transition variance matrices $\mathbf{Q}_{c,0}=[\mathbf{Q}_{c,0}^1, \mathbf{Q}_{c,0}^2, ..., \mathbf{Q}_{c,0}^{N_g}]^\top$. In practice, each initial GP covariance $\Kb_0^i$ is set to be higher than the respective planning covariance $\Kb^i$, therefore increasing likelihood of covering different homotopy classes (if they exist) while maintaining dense, local sampling for planning updates (similar to having initial exploratory behavior).

For system execution, the converged plans $\mutens$ are evaluated, and the final trajectory can be selected as the lowest-cost trajectory $\mub \in \mutens$. In the case where the method is used to gather demonstrations (as described in the main text),  this step can be ignored, and all solutions $\mutens$ stored along with contextual data $\mathbf{X}_g$ and $E(\cdot)$. 
The full algorithm for Multi-StochGPMP is shown in \cref{algo:svtrajopt_multimode}.

\begin{figure}[!t]
  \begin{subfigure}[b]{0.5\linewidth}
    \centering
    \includegraphics[height=0.75\textwidth]{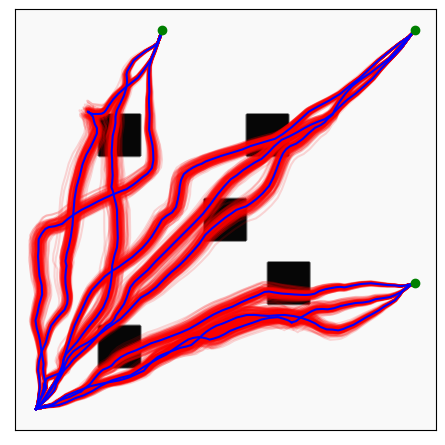}
    \caption{0 Iterations}
    \label{subfig:start}
  \end{subfigure}
  \begin{subfigure}[b]{0.5\linewidth}
    \centering
    \includegraphics[height=0.75\textwidth]{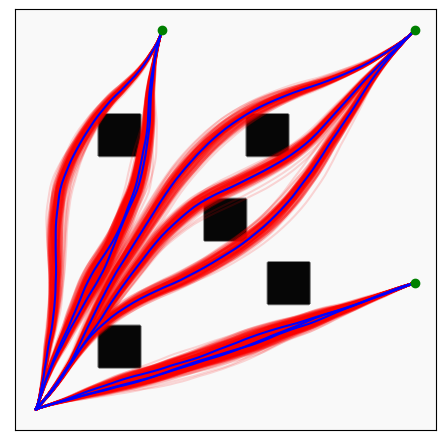}
    \caption{100 Iterations}
    \label{subfig:optimized}
  \end{subfigure}
\caption{Example of multi-StochGPMP applied to the multi-goal setting for a planar navigation problem. The GP covariance matrices are constructed using a constant-velocity stochastic transition model~\cite{barfoot2014batch}. Three goal locations (green dots) are used to initialize the mean sampling trajectories (blue), with three Gaussian process distributions per goal. Trajectory samples are generated in batch (red), resolving local collisions induced by the presence of obstacles (black). Smooth mean-trajectory solutions are generated after a few planning iterations, covering different paths to the goals.}
\label{fig:planning_example}
\end{figure}

An example of the algorithm being applied for a 2D trajectory optimization is shown in \cref{fig:planning_example}. Here, a simple point-particle linear system is defined in cartesian space, with $\textbf{Q}_c =\sigma^{-2}\,\mathbf{I}\in\mathbb{R}^{2\times 2}$. The optimization is performed over position-velocity trajectories of length $N=64$, with step size $\gamma=0.5$.

\subsection{Related Work}

The use of structured, time-correlated covariance matrices for sample-based trajectory optimization has been examined in prior work~\cite{kalakrishnan2011stomp, paraschos2013probabilistic, shyam2019improving, osa2020multimodal, kim2021derivative}. In the STOMP algorithm~\cite{kalakrishnan2011stomp}, a precision matrix with a band-diagonal structure is proposed for generating discrete-time trajectories. In comparison, the GP covariance in \cref{sec:gp_prior} allows for sampling of system velocities (or potentially higher derivatives) which are time-correlated with state according to a principled, system-based derivation of the prior distribution. Given the continuous-time representation, this also provides an option to densify trajectories through GP-interpolation. The factored form of the prior distribution (\cref{eq:gp_prior}) results in fine-grained, yet intuitive, control over the construction of the $\Kb$ matrix, \textit{i.e.} the start, goal, and transition covariances can be varied and set independently according to the desired coverage of the sampling distribution.

GP-based sampling priors have been examined for uni-modal motion planning in \cite{petrovic2020cross}. Although a similar approach is taken to \cref{algo:svtrajopt_unimode}, we take an information-theoretic perspective in deriving the update rule in \cref{sec:mp_derivation}. Here, we draw connections to existing approaches for stochastic optimal control, which only consider proposal distributions over the control space. Further, we motivate our approach towards the batch setting, where sampling and optimization can be performed in parallel. Concurrently to this work, a multi-modal planning algorithm was proposed in  \cite{petrovic2022mixtures} which uses a mixture of Gaussian process distributions. Here, sampling of trajectories is performed sequentially by first drawing from the latent variable distribution, then sampling from the resulting GP component. Given that our multi-modal, multi-goal planning method in \cref{algo:svtrajopt_multimode} is intended for batch computation using vectorized representations, sampling and distribution updates are performed simultaneously across all modes of the proposal distribution.

\section{Extended Experimental Evaluation}
\label{app:extended_exp_evaluation}

In the following, we provide an extended description of the experiments we performed. All the three experiments are divided in two parts: the learning of the \gls{ebm} components and the trajectory optimization problem using the learned \gls{ebm} as additional implicit priors.

In the three experiments, the trajectory optimization objective function was defined by a weighted sum of a set of heuristic costs, introduced in \cref{tab:costs} and the learned \gls{ebm} components. To find the optimal trajectory, we considered both sampling based methods~(Exp I and Exp II), presented in \cref{app:stoch_gpmp} and gradient based methods~(Exp III). 
The experiments were performed to validate the possibility of using \gls{ebm} as costs in trajectory optimization problems and to measure the benefit of the proposed algorithmic decisions in the performance of the obtained trajectories.

\subsection{Experiment I: 2D point Navigation}

As described in the main text, datasets for the planar navigation problem were generated by either random uniform sampling or stochastic trajectory generation with dimensions $x: \left[-10, 10\right], y: \left[-10, 10\right]$ sized region. For random sampling, 1024 example points were generated in the free-space regions of obstacles for each environment (i.e. obstacle configuration). In the case of trajectory demonstrations, for each environment, 15 random goal locations were sampled along the axis defined by $x: \left[-9, 9\right], y=9$ and $x=9, y:\left[-9, 9\right]$, with start location fixed at $x=-9,y=-9$. Trajectories were generated using the \gls{sgpmp} optimizer, with 5 trajectories of length 64 generated for each goal location. Each trajectory is selected to be the mean of each optimized \gls{sgpmp} distribution. The planning parameters were set to a timestep of $dt=0.02$, temperature $\lambda=1$, goal-factor stand-dev. $\sigma_{s}=1\times10^{-2}$, start-factor stand-dev. $\sigma_{g}=1\times10^{-2}$, GP-factor stand-dev. $\sigma_{gp}=0.1$, and obstacle-factor stand-dev. $\sigma_{obs}=1\times10^{-5}$. For sampling-based trajectory optimization, a sample size of 32 was used. A total of 512 environments were generated for each experiment. The EBM model consisted of a simple 2-hidden layer MLP (hidden width=512), with 2-D obstacle locations concatenated to the 2-dimensional inputs.

\subsection{Experiment II: Planar Manipulator}

This experiment evaluates how object-centric \gls{ebm} helps solving the task of grasp and insert object into a walled-cubby. As mentioned, we consider a simulated planar manipulator where the task space is 2D. The objective is to find a smooth trajectory in the joint space from initial joint configuration $\vq_0$ to grasp the white cube and insert the white cube into the cubby while avoiding collisions. Note that the white cube in our case is a visual mesh having no collision model.

\textbf{Dataset generation}

In this experiment, we consider a simple grasp mechanism, where the planar robot arm can grasp the object just by touching (e.g. suction, magnetic pull mechanisms).
For generating grasp points, we define four 2D uniform distributions with small margin along four edges of the cube (observing from the top view). For each side edges, we samples uniformly 10000 points and hence in total the number of positive samples are 40000. 
For generating insert trajectories, we sample uniformly 512 initial points in the planar task space. Then for each initial point, we optimize for \gls{gp} trajectory distribution having 32 time steps using our StochGPMP planner and sample 128 collision-free trajectories from the solution distribution. In total, we have $512 \times 128 \times 32$ positive samples for training insert \gls{ebm}.
While the effect number of positive samples for training planar \gls{ebm}s priors does not need to be large, we observe that the positive samples should have a suitable population to cover relevant task spaces to give informative guidance for motion optimization.

\textbf{\gls{ebm} training}

The learned \gls{ebm}s represents a state distribution

\begin{align}
p(\vx) \propto \exp \left(-E_{\vtheta}(\vx - \vx_{\textrm{frame}}) \right).
\end{align}

 with regard to their respective object-centric frame. For both grasp and insert case, the object-centric \gls{ebm}s are trained by first transforming the dataset points into the object frame of reference in consideration, then we performance training as described in \cref{sec:ebm_mo}. The network architecture used for both cases are a simple fully-connected network with two hidden layers having the width of 512.

\textbf{Trajectory Optimization with learned \gls{ebm}}

We define the manipulation planning objective with multiple cost terms: (i) \gls{gp} prior factor cost encouraging trajectory smoothness and connecting starting and final configuration, which can be computed by taking negative logarithm of \cref{eq:gp_prior}, (ii) obstacle avoidance cost and (iii) joint limit cost as specified in \cref{tab:costs}. And finally, (iv) learned grasp \gls{ebm} and (v) learned insert \gls{ebm} cost as:

\begin{align}
c(\vq) = E_{\textrm{grasp}}(FK(\vq) - \vx_{\textrm{cube}}) + E_{\textrm{insert}}(FK(\vq) - \vx_{\textrm{cubby}})
\end{align}

where $FK(\cdot)$ is forward kinematic function.

\subsection{Experiment III: Robot Pouring amid obstacles}
In this experiment, we evaluate the integration of learned \gls{ebm} components for solving a pouring task amid obstacles. We consider a 7 DoF Kuka-LWR robot manipulator. Given the robot's initial joint configuration $\vq_0$, we aim to find a trajectory that moves the robot to pour in an arbitrarily positioned pot and recover back to the initial joint configuration.

\textbf{\gls{ebm} training}

The learned \gls{ebm} represents a distribution $p(\vtau)$ that provides high probability to the trajectories performing the pouring task and low probability to the rest. For this experiment, we consider modelling this distribution with a phase-conditioned \gls{ebm}, introduced in \cref{eq:phase_ebm_trj}. Representing the whole trajectory distribution with a single \gls{ebm} might be hard due to the high dimensionality. A trajectory $\vtau$ is represented in a $N\times T$ dimensional space, with $N$ the state dimension and $T$ the temporal dimension.
Additionally, different trajectories might vary in temporal length, changing the temporal dimension $T$ of the input. To deal with these limitations, the phase-conditioned \gls{ebm}, $E_{\vtheta}(\vx, \alpha)$ represents the distribution of the state for a particular instant of the trajectory. The temporal instant is set by an $\alpha$ conditioning variable that goes from $0$ representing the start of the trajectory to $1$ the end of the trajectory. Then, the whole trajectory distribution is represented by 
\begin{align}
p(\vtau) \propto \exp \left(-\sum_{k} E_{\vtheta}(\vx_k,\alpha_k)  + (\vx_k - \vx_{k+1})^2 \right).
\end{align}
with $k$ representing different instants of the trajectory and $(\vx_k - \vx_{k+1})^2$ a smoothness cost to guarantee coherence in the trajectory. For our problem, we represent the spatial dimension $N$ in a $6$ dimensional Euclidean space, $\vx \in \RR^6$ consisting on the 3D position of the cup's bottom and the 3D position of the top of the cup. The position of this two points is represented in the pouring pot's reference frame. This way, the learned \gls{ebm} can directly adapt to arbitrary poses in the pouring pot.
Additionally, we consider as baseline an \gls{ebm} trained in the whole trajectory $p(\vtau) \propto \exp(-E_{\vtheta}(\vtau))$. We chose this baseline to evaluate the benefit of a structured distribution representation w.r.t. modelling the distribution with a single \gls{ebm}.

We train the model with \gls{cd} loss with a uniform distribution as negative sample generation. Additionally, we add a denoising score matching loss as regularizer. As previously introduced, we expect the denoising regularizer enhance the landscape of the \gls{ebm} for motion planning. We also consider as baseline an \gls{ebm} trained without the regularizer to evaluate the benefits of adding it.

\textbf{Trajectory Optimization with learned \gls{ebm}}

Given the learned pouring \gls{ebm}, we evaluate the performance benefit from adding the \gls{ebm} as additional cost function in the objective function.
To solve the pouring task, we define an objective function with multiple cost functions. A configuration space potential cost in the initial and final configuration to encourage the initial and final configuration to be a desired pose
\begin{align}
    c(\vq) = \frac{1}{2} \norm{\vq - \vq^*}^2 ,
\end{align}
with $\vq^*$ the target configuration. A table and obstacle avoidance cost, similar to the one presented in \cref{tab:costs}. A trajectory smoothness cost
\begin{align}
    c(\vq_{0:T}) = \frac{1}{T} \sum_{k=0}^{T-1}\norm{\vq_{k} - \vq_{k+1}}^2 ,
\end{align}
that encourages neighbours states to be close to each other.
A pouring avoidance cost, that encourages the orientation of the cup to be pointing up without spilling in almost all the trajectory points. To compute the cost, we compute the angle between the z axis vector in the cup $\vv_{z\_glass}$ and the z axis on the world frame $\vv_z$,  $\theta = \cos^{-1}(\vv_z \cdot \vv_{z\_glass})$. The cost aims to minimize the angle
\begin{align}
    c(\theta) = \norm{\theta}^2 .
\end{align}
Finally, a pouring cost that encourages the glass to be close to the pot and with a certain angle to pour properly. Additionally, we add a learned \gls{ebm} trained on trajectories performing the pouring task. 
In contrast with the previous experiments that apply sampling based optimization methods, we find the optimal trajectory by gradient descent. To reduce the possibility of getting a locally optimal solution, we initialize multiple particles and evolve all in batch.
The evaluation is presented in \cref{sec:experiments}.

\end{appendices}

\end{document}